\PassOptionsToPackage{table}{xcolor}
\documentclass[fleqn,10pt]{wlscirep}
\usepackage[utf8]{inputenc}
\usepackage[T1]{fontenc}
\usepackage{float}
\usepackage{caption}
\usepackage{amsmath}
\usepackage{hyperref}
\usepackage{lettrine}
\definecolor{myred}{rgb}{1.0, 0.0, 0.0}
\definecolor{myblue}{rgb}{0.0, 0.0, 1.0}
\DeclareCaptionLabelFormat{adja-page}{}

\title{\LARGE Interpretable bilinear attention network with domain adaptation improves drug-target prediction}

\author[1]{Peizhen Bai}
\author[2]{Filip Miljković}
\author[3]{Bino John}
\author[1*]{Haiping Lu}
\affil[1]{Department of Computer Science, University of Sheffield, Sheffield, United Kingdom}
\affil[2]{Imaging and Data Analytics, Clinical Pharmacology \& Safety Sciences, R\&D, AstraZeneca, Gothenburg, Sweden}
\affil[3]{Imaging and Data Analytics, Clinical Pharmacology \& Safety Sciences, R\&D, AstraZeneca, Waltham, USA}

\affil[*]{corresponding author: Haiping Lu (h.lu@sheffield.ac.uk)}

\usepackage{lineno}
\linenumbers

\begin{abstract}

Predicting drug-target interaction is key for drug discovery. Recent deep learning-based methods show promising performance but two challenges remain: (i) how to explicitly model and learn local interactions between drugs and targets for better prediction and interpretation; (ii) how to generalize prediction performance on novel drug-target pairs from different distribution. In this work, we propose DrugBAN, a deep bilinear attention network (BAN) framework with domain adaptation to explicitly learn pair-wise local interactions between drugs and targets, and adapt on out-of-distribution data. DrugBAN works on drug molecular graphs and target protein sequences to perform prediction, with conditional domain adversarial learning to align learned interaction representations across different distributions for better generalization on novel drug-target pairs. Experiments on three benchmark datasets under both in-domain and cross-domain settings show that DrugBAN achieves the best overall performance against five state-of-the-art baselines. Moreover, visualizing the learned bilinear attention map provides interpretable insights from prediction results.

\end{abstract}

\begin{document}
\nolinenumbers
\flushbottom
\maketitle
\thispagestyle{empty}
\section*{Introduction}
Drug-target interaction (DTI) prediction serves as an important step in the process of drug discovery \cite{Luo2017ANI, ztrk2018DeepDTADD, Yamanishi2008PredictionOD}. Traditional biomedical measuring from \textit{in vitro} experiments is reliable but has notably high cost and time-consuming development cycle, preventing its application on large-scale data \cite{Zitnik2019MachineLF}. In contrast, identifying high-confidence DTI pairs by \textit{in silico} approaches can greatly narrow down the search scope of compound candidates, and provide insights into the causes of potential side effects in drug combinations. Therefore, \textit{in silico} approaches have gained increasing attention and made much progress in the last few years \cite{Bagherian2021MachineLA, Wen2017DeepLearningBasedDI}.

For \textit{in silico} approaches, traditional structure-based and ligand-based virtual screening (VS) methods have been studied widely for their decent performance \cite{Sieg2019InNO}. However, structure-based VS requires molecular docking simulation, which is not applicable if the target protein's three-dimensional (3D) structure is unknown. On the other hand, ligand-based VS predicts new active molecules based on the known actives to the same protein, but the performance is poor when the number of known actives is insufficient \cite{Lim2021ARO}.

More recently, deep learning (DL)-based approaches have rapidly progressed for computational DTI prediction due to their successes in other areas, enabling large-scale validation in a relatively short time \cite{Gao2018InterpretableDT}. Many of them are constructed from a chemogenomics perspective \cite{Bredel2004ChemogenomicsAE, Yamanishi2008PredictionOD}, which integrates the chemical space, genomic space, and interaction information into a unified end-to-end framework. Since the number of biological targets that have available 3D structures is limited, many DL-based models take linear or two-dimensional (2D) structural information of drugs and proteins as inputs. They treat DTI prediction as a binary classification task, and make predictions by feeding the inputs into different deep encoding and decoding modules such as deep neural network (DNN) \cite{Lee2019DeepConvDTIPO, Hinnerichs2021DTIVoodooML}, graph neural network (GNN) \cite{Gao2018InterpretableDT, Nguyen2020GraphDTAPD, Tsubaki2019CompoundproteinIP, feng2018padme} or transformer architectures \cite{Chen2020TransformerCPIIC, Huang2021MolTransMI}. With the advances of deep learning techniques, such models can automatically learn data-driven representations of drugs and proteins from large-scale DTI data instead of only using pre-defined descriptors.

Despite these promising developments, two challenges remain in existing DL-based methods. The first challenge is explicit learning of interactions between local structures of drug and protein. DTI is essentially decided by mutual effects between important molecular substructures in the drug compound and binding sites in the protein sequence  \cite{Schenone2013TargetIA}. However, many previous studies learn global representations in their separate encoders, without explicitly learning local interactions \cite{ztrk2018DeepDTADD, Nguyen2020GraphDTAPD, Ozturk2019WideDTAPO, Zheng2020PredictingDI, Lee2019DeepConvDTIPO}. Consequently, drug and protein representations are learned for the whole structures first and mutual information is only implicitly learned in the black-box decoding module. Interactions between drug and target are particularly related to their crucial substructures, thus separate global representation learning tends to limit the modeling capacity and prediction performance. Moreover, without explicit learning of local interactions, the prediction result is hard to interpret, even if the prediction is accurate. 

\begin{figure*}[t!]
    \begin{center}
    \includegraphics[width=1\textwidth]{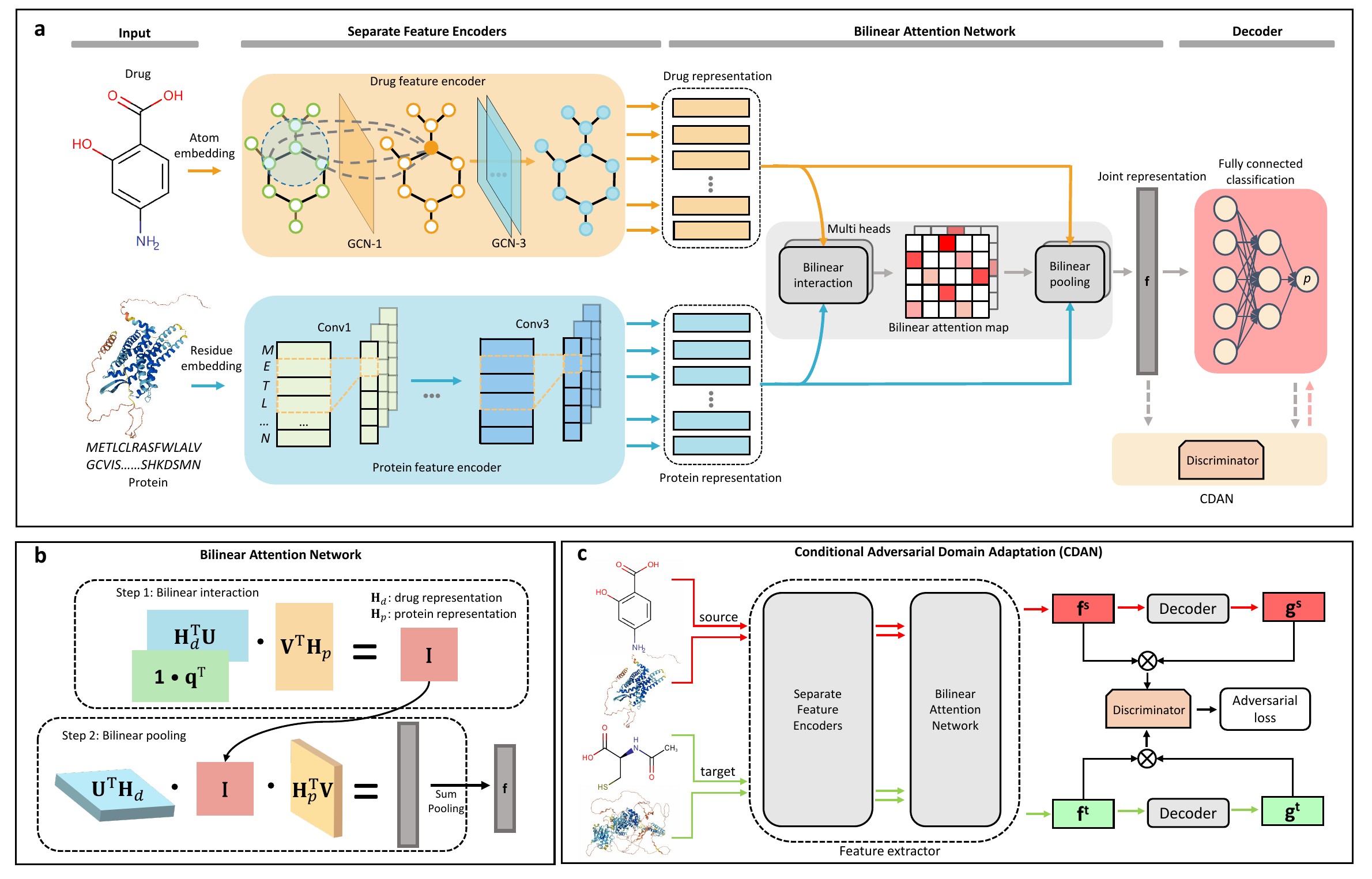}
    \end{center}
    \caption{\textbf{Overview of the DrugBAN framework.} \textbf{(a)} The input drug molecule and protein sequence are separately encoded by graph convolutional networks and 1D-convolutional neural networks. Each row of the encoded drug representation is an aggregated representation of adjacent atoms in the drug molecule, and each row of the encoded protein representation is a subsequence representation in the protein sequence. The drug and protein representations are fed into a bilinear attention network to learn their pairwise local interactions. The joint representation $\mathbf{f}$ is decoded by a fully connected decoder module to predict the DTI probability $p$. If the prediction task is cross-domain, the conditional domain adversarial network \cite{Long2018ConditionalAD} (CDAN) module is employed to align learned representations in the source and target domains. \textbf{(b)} The bilinear attention network architecture. $\mathbf{H}_d$ and $\mathbf{H}_p$ are encoded drug and protein representations. In Step 1, the bilinear attention map matrix $\mathbf{I}$ is obtained by a low-rank bilinear interaction modeling via transformation matrices $\mathbf{U}$ and $\mathbf{V}$ to measure the substructure-level interaction intensity \cite{Kim2017}. Then $\mathbf{I}$ is utilized to produce the joint representation $\mathbf{f}$ in Step 2 by bilinear pooling via the shared transformation matrices $\mathbf{U}$ and $\mathbf{V}$. \textbf{(c)} CDAN is a domain adaptation technique to reduce the domain shift between different distributions of data. We use CDAN to embed joint representation $\mathbf{f}$ and softmax logits $\mathbf{g}$ for source and target domains into a joint conditional representation via the discriminator, a two-layer fully connected network that minimizes the domain classification error to distinguish the target domain from the source domain.}
    \label{fig:workflow}
\end{figure*}

The second challenge is generalizing prediction performance across domains, i.e. out of learned distribution. Due to the vast regions of chemical and genomic space, drug-target pairs that need to be predicted in real-world applications are often unseen and dissimilar to any pairs in the training data. They have different distributions and thus need cross-domain modeling \cite{Abbasi2020DeepCDADC, kao2021toward}. A robust model should be able to transfer learned knowledge to a new domain that only has unlabeled data. In this case, we need to align distributions and improve cross-domain generalization performance by learning transferable representations, e.g. from "source" to "target". To the best of our knowledge, this is an underexplored direction in drug discovery \cite{abbasi2021deep}.

To address these challenges, we propose an interpretable bilinear attention network-based model (DrugBAN) for DTI prediction, as shown in Figure \ref{fig:workflow}a. DrugBAN is a deep learning framework with explicit learning of local interactions between drug and target, and conditional domain adaptation for learning transferable representations across domains. Specifically, we first use graph convolutional network \cite{Kipf2017SemiSupervisedCW} (GCN) \cite{} and convolutional neural network (CNN) to encode local structures in 2D drug molecular graph and 1D protein sequence, respectively. Then the encoded local representations are fed into a pairwise interaction module that consists of a bilinear attention network \cite{Yu2018BeyondBG, Kim2018BilinearAN} to learn local interaction representations, as depicted in Figure \ref{fig:workflow}b. The local joint interaction representations are decoded by a fully connected layer to make a DTI prediction. In this way, we can utilize the pairwise bilinear attention map to visualize the contribution of each substructure to the final predictive result, improving the interpretability. For cross-domain prediction, we apply conditional domain adversarial network \cite{Long2018ConditionalAD} (CDAN) to transfer learned knowledge from source domain to target domain to enhance cross-domain generalization, as illustrated in Figure \ref{fig:workflow}c. We conduct a comprehensive performance comparison against five state-of-the-art DTI prediction methods on both in-domain and cross-domain settings of drug discovery. The results show that our method achieves the best overall performance compared to state-of-the-art methods, while providing interpretable insights for the prediction results.

To summarize, DrugBAN differs from previous works by (i) capturing pairwise local interactions between drugs and targets via a bilinear attention mechanism, (ii) enhancing cross-domain generalization with an adversarial domain adaptation approach; and (iii) giving an interpretable prediction via bilinear attention weights instead of black-box results.

\section*{Results}
\subsection*{Problem formulation}
In DTI prediction, the task is to determine whether a pair of a drug compound and a target protein will interact. For target protein, denoting each protein sequence as $\mathcal{P} = (a_1,...,a_n)$, where each token $a_i$ represents one of the 23 amino acids. For drug compound, most existing deep learning-based methods represent the input by the Simplified Molecular Input Line Entry System (SMILES) \cite{Weininger1988SMILESAC}, which is a 1D sequence describing chemical atom and bond token information in the drug molecule. The SMILES format allows encoding drug information with many classic deep learning architectures. However, since the 1D sequence is not a natural representation for molecules, some important structural information of drugs could be lost, degrading model prediction performance. Our model converts input SMILES into its corresponding 2D molecular graph. Specifically, a drug molecule graph is defined as $\mathcal{G} = (\mathcal{V}, \mathcal{E})$, where $\mathcal{V}$ is the set of vertices (atoms) and $\mathcal{E}$ is the set of edges (chemical bonds).

Given a protein sequence $\mathcal{P}$ and a drug molecular graph $\mathcal{G}$, DTI prediction aims to learn a model $\mathcal{M}$ to map the joint feature representation space $\mathcal{P} \times \mathcal{G}$ to an interaction probability score $p \in [0, 1]$. Supplementary Table 3 provides the commonly used notations in this paper.

\subsection*{DrugBAN framework}
Figure \ref{fig:workflow}a shows the proposed DrugBAN framework. Given an input drug-target pair, firstly, we employ separate graph convolutional network (GCN) and 1D-convolutional neural network (1D-CNN) blocks to encode molecular graph and protein sequence information, respectively. Then we use a bilinear attention network module to learn local interactions between encoded drug and protein representations. The bilinear attention network consists of a bilinear attention step and a bilinear pooling step to generate a joint representation, as illustrated in Figure \ref{fig:workflow}b. Next, a fully connected classification layer learns a predictive score indicating the probability of interaction. For improving model generalization performance on cross-domain drug-target pairs, we further embed CDAN into the framework to adapt representations for better aligning source and target distributions, as depicted in Figure \ref{fig:workflow}c. 

\subsection*{Evaluation strategies and metrics}
We study classification performance on three public datasets separately: BindingDB \cite{Liu2007BindingDBAW}, BioSNAP \cite{biosnapnets} and Human \cite{Liu2015ImprovingCI, Chen2020TransformerCPIIC}, with test sets holding out as `unknown' for evaluation. We use two different split strategies for in-domain and cross-domain settings. For in-domain evaluation, each experimental dataset is randomly divided into training, validation, and test sets with a 7:1:2 ratio. For cross-domain evaluation, we propose a clustering-based pair split strategy to construct cross-domain scenario. We conduct cross-domain evaluation on the large-scale BindingDB and BioSNAP datasets. For each dataset, we firstly use the single-linkage algorithm to cluster drugs and proteins by ECFP4 (extended connectivity fingerprint, up to four bonds) \cite{Rogers2010ExtendedConnectivityF} fingerprint and pseudo amino acid composition (PSC) \cite{Cao2013propyAT}, respectively. After that, we randomly select 60\% drug clusters and 60\% protein clusters from the clustering result, and consider all drug-target pairs between the selected drugs and proteins as source domain data. All the  pairs between drugs and proteins in the remaining clusters are considered to be target domain data. The clustering implementation details are provided in Supplementary Section 1. Under the clustering-based pair split strategy, the source and target domains are non-overlapping with different distributions. Following the general setting of domain adaptation, we use all labeled source domain data and 80\% unlabeled target domain data as the training set, and the remaining 20\% labeled target domain data as the test set. The cross-domain evaluation is more challenging than in-domain random split but provides a better measure of model generalization ability in real-world drug discovery. For a more comprehensive study, we report additional experiments across different protein families, on unseen drugs/targets, and with high fraction of missing data in Supplementary Sections 4-6, respectively.

The AUROC (area under the receiver operating characteristic curve) and AUPRC (area under the precision-call curve) are used as the major metrics to evaluate model classification performance. In addition, we also report the accuracy, sensitivity, and specificity at the threshold of the best F1 score. We conduct five independent runs with different random seeds for each dataset split. The best performing model is selected to be the one with the best AUROC on the validation set. The selected model is then evaluated on the test set to report the performance metrics.

\begin{table*}[t]
\centering
\small
\caption{\centering In-domain performance comparison on the BindingDB and BioSNAP datasets with random split (\textbf{Best}, \underline{Second Best}).}
\setlength{\tabcolsep}{5mm}{\begin{tabular}{llllll}
\toprule
Method                 & AUROC & AUPRC & Accuracy &Sensitivity & Specificity \\ \midrule
\multicolumn{6}{c}{BindingDB} \\
SVM \cite{cortes1995support} & 0.939$\pm$0.001 & 0.928$\pm$0.002 & 0.825$\pm$0.004 & 0.781$\pm$0.014 & 0.886$\pm$0.012 \\
RF \cite{ho1995random}       & 0.942$\pm$0.011        & 0.921$\pm$0.016   & 0.880$\pm$0.012    & 0.875$\pm$0.023            & 0.892$\pm$0.020  \\
DeepConv-DTI \cite{Lee2019DeepConvDTIPO}           & 0.945$\pm$0.002       & 0.925$\pm$0.005 &0.882$\pm$0.007       & 0.873$\pm$0.018            & 0.894$\pm$0.009 \\
GraphDTA \cite{Nguyen2020GraphDTAPD}      & 0.951$\pm$0.002        & 0.934$\pm$0.002 & \underline{0.888$\pm$0.005}       & \underline{0.882$\pm$0.012}            & 0.897$\pm$0.008 \\
MolTrans \cite{Huang2021MolTransMI} & \underline{0.952$\pm$0.002}        & \underline{0.936$\pm$0.001} & 0.887$\pm$0.006     & 0.877$\pm$0.016        & \underline{0.902$\pm$0.009} \\
\rowcolor{gray!10}
DrugBAN & \textbf{0.960$\pm$0.001}        & \textbf{0.948$\pm$0.002} & \textbf{0.904$\pm$0.004}       & \textbf{0.900$\pm$0.008} & \textbf{0.908$\pm$0.004} \\
\multicolumn{6}{c}{BioSNAP} \\
SVM \cite{cortes1995support}             & 0.862$\pm$0.007       & 0.864$\pm$0.004  & 0.777$\pm$0.011        & 0.711$\pm$0.042  & 0.841$\pm$0.028   \\
RF \cite{ho1995random}         & 0.860$\pm$0.005       & 0.886$\pm$0.005  & 0.804$\pm$0.005        & \textbf{0.823$\pm$0.032}            & 0.786$\pm$0.025   \\
DeepConv-DTI \cite{Lee2019DeepConvDTIPO}           & 0.886$\pm$0.006 & 0.890$\pm$0.006  & 0.805$\pm$0.009        & 0.760$\pm$0.029            & \underline{0.851$\pm$0.013}   \\
GraphDTA \cite{Nguyen2020GraphDTAPD}      & 0.887$\pm$0.008       & 0.890$\pm$0.007  & 0.800$\pm$0.007        & 0.745$\pm$0.032            & \textbf{0.854$\pm$0.025}   \\
MolTrans \cite{Huang2021MolTransMI} & \underline{0.895$\pm$0.004}       & \underline{0.897$\pm$0.005}  & \underline{0.825$\pm$0.010}        & 0.818$\pm$0.031            & 0.831$\pm$0.013   \\
\rowcolor{gray!10}
DrugBAN & \textbf{0.903$\pm$0.005}       & \textbf{0.902$\pm$0.004}  & \textbf{0.834$\pm$0.008}        & \underline{0.820$\pm$0.021}            & 0.847$\pm$0.010   \\
\bottomrule
\end{tabular}}
\label{random_split}
\end{table*}

\begin{figure}[tbp]
\centering
\hspace*{-0.65cm}  
    \includegraphics[width=0.7\textwidth]{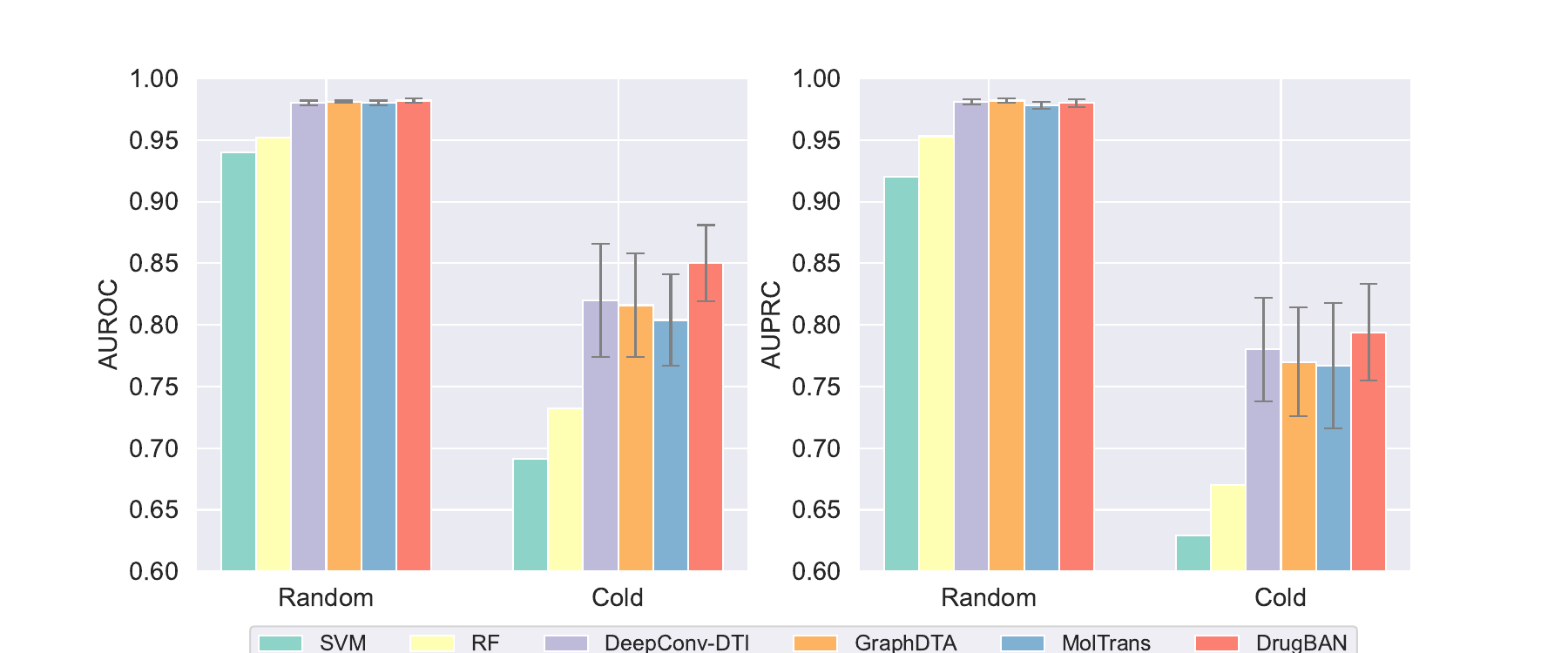}
    \caption{\textbf{In-domain performance comparison on the Human dataset with random split and cold pair split.} The grey lines are error bars indicating the standard deviation.}
    \label{fig:human}
\end{figure}

\subsection*{In-domain performance comparison}
Here we compare DrugBAN with five baselines under the random split setting: support vector machine \cite{cortes1995support} (SVM), random forest \cite{ho1995random} (RF), DeepConv-DTI \cite{Lee2019DeepConvDTIPO}, GraphDTA \cite{Nguyen2020GraphDTAPD}, and MolTrans \cite{Huang2021MolTransMI}. This is the in-domain scenario so we use vanilla DrugBAN without embedding the CDAN module. Table \ref{random_split} shows the comparison on the BindingDB and BioSNAP datasets. DrugBAN has consistently outperformed baselines in AUROC, AUPRC, and accuracy, while the performance in sensitivity and specificity is also competitive. The results indicate that data-driven representation learning can capture more important information than pre-defined descriptor features in in-domain DTI prediction. Moreover, DrugBAN can capture interaction patterns via its pairwise interaction module, further improving prediction performance.

Figure \ref{fig:human} shows the in-domain results on the Human dataset. Under the random split, the deep learning-based models all achieve similar and promising performance (AUROC $>$ 0.98). However, Chen et al. (2020) \cite{Chen2020TransformerCPIIC} pointed out that the Human dataset had some hidden ligand bias, resulting in the correct predictions being made only based on the drug features rather than interaction patterns. The high accuracy could be due to bias and overfitting, not indicating a model's real-world performance on prospective prediction. Therefore, we further use a cold pair split strategy to evaluate models to mitigate the overoptimism of performance estimation under random split due to the data bias. This cold pair split strategy guarantees that all test drugs and proteins are not observed during training so that prediction on test data cannot rely only on the features of known drugs or proteins. We randomly assign 5\% and 10\% DTI pairs into the validation and test sets respectively, and remove all their associated drugs and proteins from the training set. Figure \ref{fig:human} indicates that all models have a significant performance drop from random split to cold pair split, especially for SVM and RF. However, we can see that DrugBAN still achieves the best performance against other state-of-the-art deep learning baselines.

\begin{figure*}[t!]
    \begin{center}
    \includegraphics[width=1\textwidth]{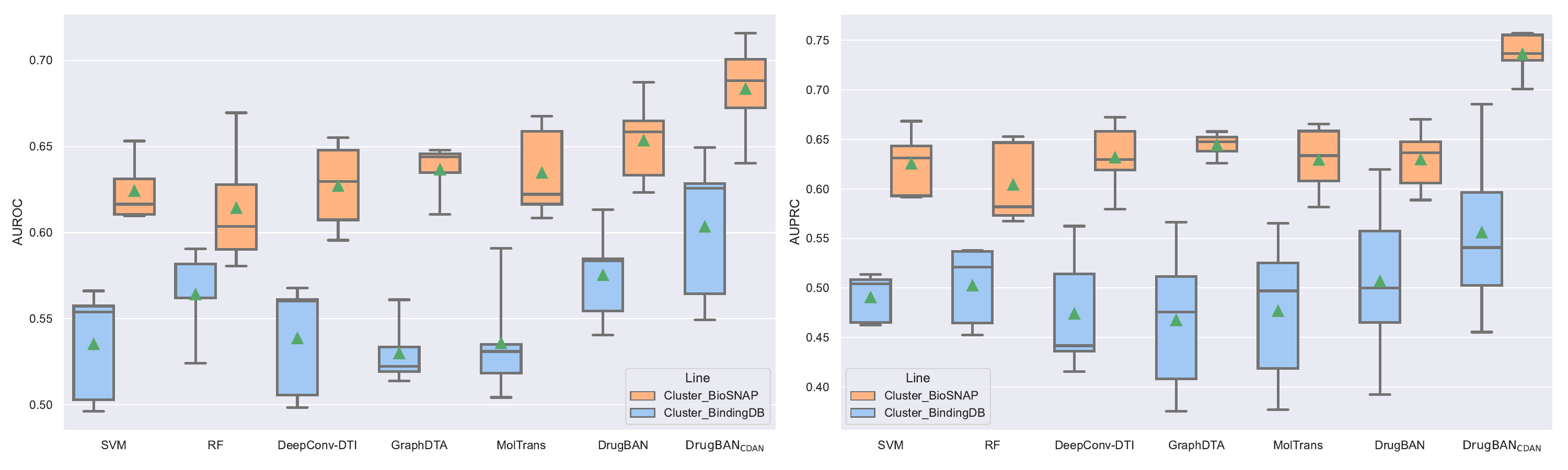}
    \end{center}
    \vspace{-2em}
    \caption{\textbf{Cross-domain performance comparison on the BindingDB and BioSNAP datasets with clustering-based pair split.} The box plots show the median as the center lines, and the mean as the green triangles.}
    \label{fig:cluster}
\end{figure*}

\begin{table}[t]
\centering
\caption{Ablation study in AUROC on the BindingDB and BioSNAP datasets with random and clustering-based split strategies (averaged over five random runs). The first four models show the effectiveness of our bilinear attention module, and the last three models show the strength of $\mathrm{DrugBAN_{CDAN}}$ on cross-domain prediction  (\textbf{Best}, \underline{Second Best}).}

\begin{tabular}{lcccc}
\toprule
Ablation tests & $\mathrm{BindingDB_{random}}$ & $\mathrm{BioSNAP_{random}}$ & $\mathrm{BindingDB_{cluster}}$ & $\mathrm{BioSNAP_{cluster}}$ \\ \midrule
Linear concatenation \cite{Lee2019DeepConvDTIPO, Nguyen2020GraphDTAPD, ztrk2018DeepDTADD} &  0.949$\pm$0.002 & 0.887$\pm$0.007 & - & - \\
One-side target attention \cite{Tsubaki2019CompoundproteinIP}  & 0.950$\pm$0.002 &  0.890$\pm$0.005 & - & - \\
One-side drug attention \cite{Tsubaki2019CompoundproteinIP}     & \underline{0.953$\pm$0.002} & \underline{0.892$\pm$0.004} & - & -\\
\rowcolor{gray!10}
DrugBAN & \textbf{0.960$\pm$0.001} & \textbf{0.903$\pm$0.005} & 0.575$\pm$0.025 & 0.654$\pm$0.023\\
$\mathrm{MolTrans_{CDAN}}$ & -       & - & 0.575$\pm$0.038 & 0.656$\pm$0.028 \\
$\mathrm{DrugBAN_{DANN}}$ & -        & - & \underline{0.592$\pm$0.042} & \underline{0.667$\pm$0.030} \\
\rowcolor{gray!10}
$\mathrm{DrugBAN_{CDAN}}$ & -        & - & \textbf{0.604$\pm$0.039}  & \textbf{0.684$\pm$0.026}  \\
\bottomrule
\end{tabular}
\label{ablation}
\end{table}

\subsection*{Cross-domain performance comparison}
In-domain classification under random split is an easier task and of less practical importance. Therefore, next, we study more realistic and challenging cross-domain DTI prediction, where training data and test data have different distributions. To imitate this scenario, the original data is divided into source and target domains by the clustering-based pair split. We turn on the CDAN module of DrugBAN to get $\mathrm{DrugBAN_{CDAN}}$ for studying knowledge transferability in cross-domain prediction.

Figure \ref{fig:cluster} presents the performance evaluation on the BindingDB and BioSNAP datasets with clustering-based pair split. Compared to the previous in-domain prediction results, the performance of all DTI models drops significantly due to much less information overlap between training and test data. In this scenario, vanilla DrugBAN still outperforms other state-of-the-art models on the whole. Specifically, it outperforms MolTrans by 2.9\% and 7.4\% in AUROC on the BioSNAP and BindingDB datasets, respectively. The results show that DrugBAN is a robust method under both in-domain and cross-domain settings. Interestingly, RF achieves good performance and even consistently outperforms other deep learning baselines (DeepConv, GraphDTA and MolTrans) on the BindingDB dataset. The results indicate that deep learning methods are not always superior to shallow machine learning methods under the cross-domain setting.

Recently, domain adaptation techniques have received increasing attention due to the ability of transferring knowledge across domains, but they are mainly applied to computer vision and natural language processing problems. We combine vanilla DrugBAN with CDAN to tackle cross-domain DTI prediction. As shown in Figure \ref{fig:cluster}, $\mathrm{DrugBAN_{CDAN}}$ has significant performance improvements with the introduction of a domain adaptation module. On the BioSNAP dataset, it outperforms vanilla DrugBAN by 4.6\% and 16.9\% in AUROC and AUPRC, respectively. By minimizing the distribution discrepancy across domains, CDAN can effectively enhance DrugBAN generalization ability and provide more reliable results. 

These results demonstrate the strength of DrugBAN in generalizing prediction performance across domains.

\subsection*{Ablation study}
Here we conduct an ablation study to investigate the influences of bilinear attention and domain adaptation modules on DrugBAN. The results are shown in Table \ref{ablation}. To validate the effectiveness of bilinear attention, we study three variants of DrugBAN that differ in the joint representation computation between drug and protein: one-side drug attention, one-side protein attention, and linear concatenation. The one-side attention is equivalent to the neural attention mechanism introduced by Tsubaki et al. (2019) \cite{Tsubaki2019CompoundproteinIP}, which is used to capture the joint representation between a drug vector representation and a protein subsequence matrix representation. We replace the bilinear attention in DrugBAN with one-side attention to generate the two variants. Linear concatenation is a simple vector concatenation of drug and protein vector representations after a max-pooling layer. As shown in the first four rows of Table \ref{ablation}, the results demonstrate that bilinear attention is the most effective method to capture interaction information for DTI prediction. To examine the effect of CDAN, we study two variants: DrugBAN with domain-adversarial neural network (DANN) \cite{Ganin2016DomainAdversarialTO} (i.e. $\mathrm{DrugBAN_{DANN}}$) and MolTrans with CDAN (i.e. $\mathrm{MolTrans_{CDAN}}$). DANN is another adversarial domain adaptation technique without considering classification distribution. The last four rows of Table \ref{ablation} indicate that $\mathrm{DrugBAN_{CDAN}}$ still achieves the best performance improvement in cross-domain prediction.

\begin{figure*}[t!]
    \begin{center}
    \includegraphics[width=1\textwidth]{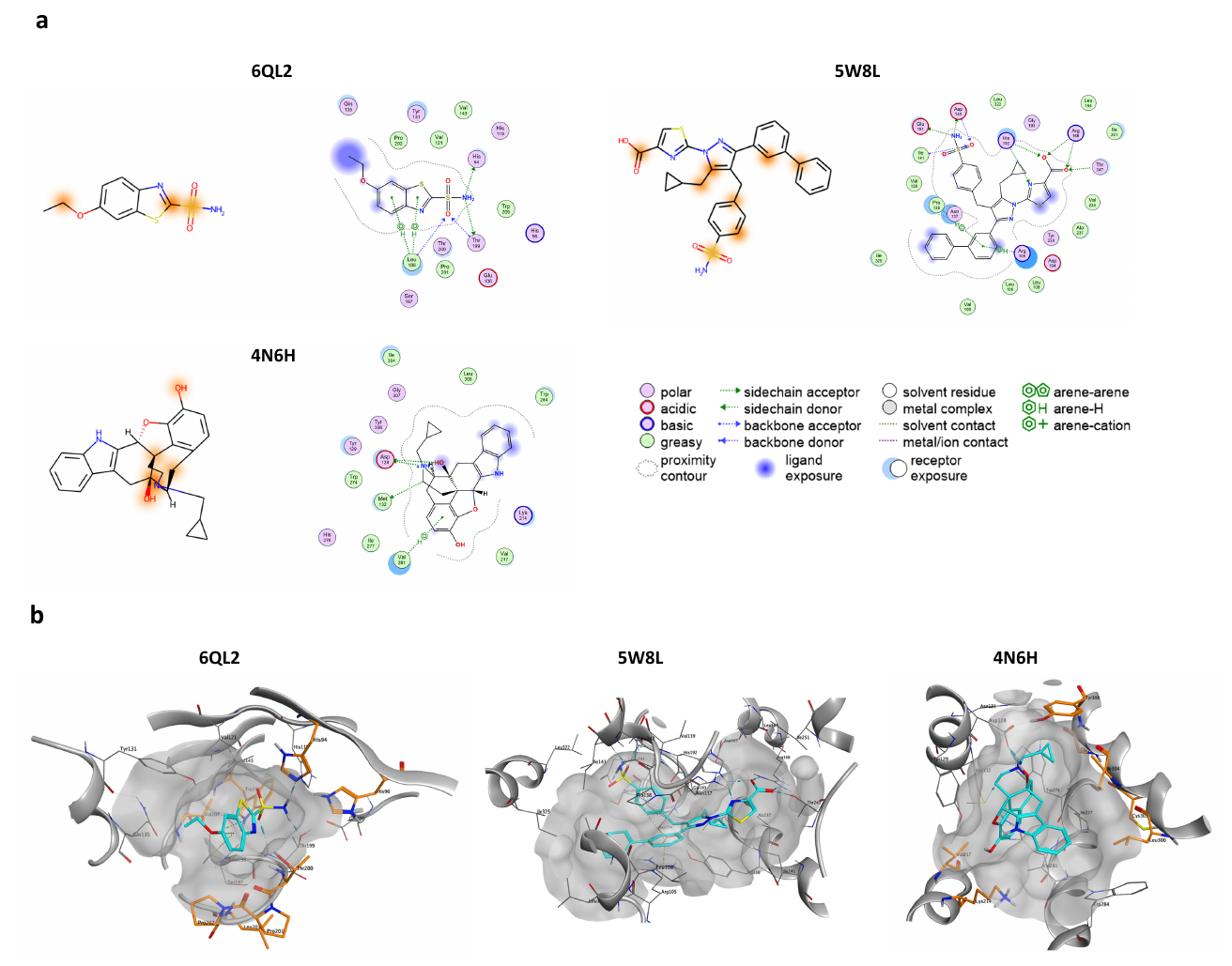}
    \end{center}
    \vspace{-2em}
    \caption{\textbf{Importance visualization of ligands and binding pockets.} \textbf{(a)} Interpretability of co-crystalized ligands. The left-hand side of each panel shows the two-dimensional structures of ligands with highlighted atoms (orange) that were predicted to contribute to protein binding. All structures were visualized using RDKit \cite{rdk}. In addition, ligand-protein interaction maps (right-hand side of each panel) from the corresponding crystal structures of these ligands are provided. At the right bottom, the legend panel for the ligand-protein interaction maps is displayed. \textbf{(b)} Interpretability of binding pocket structures. The three-dimensional representations of ligand-protein binding pockets are provided highlighting the  correctly predicted amino acid residues (orange) that surround the corresponding ligands (cyan). Remaining amino acid residues, secondary structure elements, and surface maps are colored in grey. All ligand-protein interaction maps and three-dimensional representations of X-ray structures were visualized using the Molecular Operating Environment (MOE) software \cite{moe}. See text below for interesting interpretation.}
    \label{fig:mol}
\end{figure*}

\subsection*{Interpretability with bilinear attention visualization}
A further strength of DrugBAN is to enable molecular level insights and interpretation critical for drug design efforts, utilizing the components of the bilinear attention map to visualize the contribution of each substructure to the final predictive result. Here, we examine the top three predictions (PDB ID: 6QL2 \cite{kazokaite2019engineered}, 5W8L \cite{rai2017discovery} and 4N6H \cite{fenalti2014molecular}) of co-crystalized ligands from Protein Data Bank (PDB) \cite{berman2000protein}. Only X-ray structures with resolution greater than 2.5 Å that corresponded to human protein targets were proceeded for selection. In addition, co-crystalized ligands were required to have pIC$_{50}$ $\le$ 100 nM and not to be part of the training set. The visualization results are shown in Figure \ref{fig:mol}a alongside the ligand-protein interaction maps originating from the corresponding X-ray structures. For each molecule, we colored its top 20\% weighted atoms in bilinear attention map with orange. 

For PDB structure 6QL2 (ethoxzolamide complexed with human carbonic anhydrase 2), our model correctly interpreted sulfonamide region as essential for ligand-protein binding (in 6QL2: sulfonamide oxygen as a hydrogen bond acceptor to the backbone of Leu198 and Thr199, and amino group as a hydrogen bond donor to the side chains of His94 and Thr199). On another hand, ethoxy group of ethoxzolamide was incorrectly predicted to form specific interactions with the protein, although its exposure to the solvent may promote further binding (blue highlight). In addition, benzothiazole scaffold, which forms an arene-H interaction with Leu198, is only partly highlighted by our interpretability model. It is worth mentioning that though top 20\% of interacting atoms of ethoxzolamide only corresponded to three highlighted atoms, all of them indicated different ligand-protein interaction sites corroborated by the X-ray structure. 

In 5W8L structure (9YA ligand bound to human L-lactate dehydrogenase A), the interpretability feature once more highlighted important interaction patterns for ligand-protein binding. For example, sulfonamide group was once more indicated to form specific interactions with the protein (in 5W8L: amino group as a hydrogen bond donor to the side chains of Asp140 and Glu191, and sulfonamide oxygen as a hydrogen bond acceptor to the backbone of Asp140 and Ile141). Similarly, we noted that carboxylic acid group was also partly highlighted (in 5W8L: carboxylic acid oxygens act as hydrogen bond acceptors to the side chains of Arg168, His192, and Thr247). Moreover, biphenyl rings were correctly predicted to participate in ligand-protein binding (in 5W8L: arene-H interaction with Arg105 and Asn137). Although 9YA (bound to 5W8L) was much larger and complex than ethoxzolamide (bound to 6QL2), the model showed good interpretability potential for the majority of the experimentally confirmed interactions. 

In the third example, 4N6H X-ray complex of human delta-type opioid receptor with EJ4 ligand, main interacting functional groups of EJ4 were once more highlighted correctly. Here, a hydroxyl group of the aliphatic ring complex and a neighboring tertiary amine (in 4N6H: both as hydrogen bond donors to the side chain of Asp128) were correctly interpreted to form specific interactions. On the other hand, phenol group was wrongly predicted to participate in protein binding.

As for the more challenging protein sequence interpretability, the results were overall weaker than those for the ligand interpretability. Although many amino acid residues that were predicted to potentially participate in ligand binding were in fact distantly located to the respective compounds, a number of amino acid residues forming the binding sites were yet correctly predicted, which is shown in Figure \ref{fig:mol}b. For example, in 6QL2 complex the following residues were highlighted: His94, His96, Thr200, Pro201, Pro202, Leu203, Val207, Trp209. Among these, only His94 forms specific interaction with ethoxzolamide. In 5W8L, none of the residues that constitute the ligand-protein binding site were highlighted. However, in 4N6H structure, there were several correctly predicted residues within the binding site: Lys214, Val217, Leu300, Cys303, Ile304, Gly307, and Tyr308. Unfortunately, none of the residues participated in the specific interactions with the ligand. Given these results, it is expected that protein sequence interpretability would be less confident because the one-dimensional protein sequence (used as protein information input in our model) does not necessarily imply the three-dimensional configuration and locality of the binding pocket. However, the results from the primary protein sequence are encouraging enough to safely assume that the further incorporation of three-dimensional protein information into the modeling framework would eventually improve the model interpretability of drug-target interaction networks. 

In addition, as the interpretability provided by DrugBAN is adaptively learned from DTI data itself, such interpretation has potential to find some hidden knowledge of local interactions that has not been explored, and could help drug hunters to improve binding properties of a given scaffold, or to reduce the off-target liabilities of a compound.

\section*{Conclusion}
In this work, we present DrugBAN, an end-to-end bilinear attention deep learning framework for DTI prediction. We have integrated CDAN, an adversarial domain adaptation network, into the modeling process to enhance cross-domain generalization ability. Compared with other state-of-the-art DTI models and conventional machine learning models, the experimental results show that DrugBAN consistently achieves improved DTI prediction performance in both in-domain and cross-domain settings. Furthermore, by mapping attention weights to protein subsequences and drug compound atoms, our model can provide biological insights for interpreting the nature of interactions. The proposed ideas are general in nature and can be extended to other interaction prediction problems, such as the prediction of drug-drug interaction and protein-protein interaction. 

This work focuses on chemogenomics-based DTI using 1D protein sequence and 2D molecular graph as input. Since the number of highly accurate 3D structured proteins only accounts for a small fraction of the known protein sequences, this work did not consider the modeling with such structural information. Nevertheless, DeepMind's AlphaFold \cite{Jumper2021HighlyAP} is making great progress in protein 3D structure prediction, recently generating 2 billion protein 3D structure predictions from 1 million species. Such progress opens doors for utilizing 3D structural information in chemogenomics-based DTI prediction. Following the idea of pairwise local interaction learning and domain adaptation, we believe that extending our ideas further on complex 3D structures can lead to even better performance and interpretability in future work. Finally, this work studies different datasets separately, combining dataset integration with DrugBAN will be another interesting future direction to explore.

\section*{Methods}

\subsection*{Bilinear attention network}
This is an attention-based model and was first proposed to solve the problem of visual question answering (VQA) \cite{Kim2018BilinearAN}. Given an image and relevant natural language question, VQA systems aim to provide a text-image matching answer. Therefore, VQA can be viewed as a multimodal learning task, similar to DTI prediction. Bilinear attention network (BAN) uses a bilinear attention map to gracefully extend unitary attention networks for adapting multimodal learning, which considers every pair of multimodal input channels, i.e., the pairs of image regions and question words to learn an interaction representation. Compared to using a unitary attention mechanism directly on multimodal data, BAN can provide richer joint information but keep the computational cost at the same scale. Due to the problem similarity between VQA and DTI, we design a BAN-inspired pairwise interaction module for DTI prediction.

\subsection*{Domain adaptation} These approaches learn a model that reduces domain distribution shift between the source domain and target domain, which is mainly developed and studied in computer vision \cite{Pan2010ASO}. Early domain adaptation methods tended to reweight sample importance or learn invariant feature representations in shallow feature space, using labeled data in the source domain and unlabeled data in the target domain. More recently, deep domain adaptation methods embed the adaptation module in various deep architectures to learn transferable representations \cite{Gong2013ConnectingTD, Huang2006CorrectingSS}. In particular, Long et al. (2018) \cite{Long2018ConditionalAD} proposed a novel deep domain adaptation method, CDAN, that combines adversarial networks with multilinear conditioning for transferable representation learning. By introducing classifier prediction information into adversarial learning, CDAN can effectively align data distributions in different domains. We embed CDAN as an adaptation module in DrugBAN to enhance model performance for cross-domain DTI prediction.

\subsection*{DrugBAN architecture}
\textbf{CNN for protein sequence.} The protein feature encoder consists of three consecutive 1D-convolutional layers, which transforms an input protein sequence to a matrix representation in the latent feature space. Each row of the matrix denotes a subsequence representation in the protein. Drawing on the concept of word embedding, we first initialize all amino acids into a learnable embedding matrix $\mathbf{E}_{p} \in \mathbb{R}^{23 \times D_{p}}$, where 23 is the number of amino acid types and $D_{p}$ is the latent space dimensionality. By looking up $\mathbf{E}_{p}$, each protein sequence $\mathcal{P}$ can be initialized to corresponding feature matrix $\mathbf{X}_p \in \mathbb{R}^{\Theta_{p} \times D_{p}}$. Here $\Theta_{p}$ is the maximum allowed length of a protein sequence, which is set to align different protein lengths and make batch training. Following previous works \cite{Huang2021MolTransMI, ztrk2018DeepDTADD, Tsubaki2019CompoundproteinIP}, protein sequences with maximum allowed length are cut, and those with smaller length are padded with zeros.

The CNN-block protein encoder extracts local residue patterns from the protein feature matrix $\mathbf{X}_p$. Here a protein sequence is considered as an overlapping 3-mer amino acids such as ``METLCL...DSMN'' $\rightarrow$ ``MET'', ``ETL'', ``TLC'',..., ``DSM'', ``DLK''. The first convolutional layer is utilized to capture the 3-mer residue-level features with kernel size $=3$. Then the next two layers continue to enlarge the receptive field and learn more abstract features of local protein fragments. The protein encoder is described as follows:

\begin{equation}
    \mathbf{H}_p^{(l+1)} = \sigma(\mathrm{CNN}(\mathbf{W}_{\emph{c}}^{(l)}, \mathbf{b}_{\emph{c}}^{(l)}, \mathbf{H}_p^{(l)})),
\end{equation}
where $\mathbf{W}_{\emph{c}}^{(l)}$ and $\mathbf{b}_{\emph{c}}^{(l)}$ are the learnable weight matrices (filters) and bias vector in the $l$-th CNN layer. $\mathbf{H}_p^{(l)}$ is the $l$-th hidden protein representation and $\mathbf{H}_p^{(0)} = \mathbf{X}_p$. $\sigma(\cdot)$ denotes a non-linear activation function, with $\mathrm{ReLU}(\cdot)$ used in our experiments.

\textbf{GCN for molecular graph.} For drug compound, we convert each SMILES string to its 2D molecular graph $\mathcal{G}$. To represent node information in $\mathcal{G}$, we first initialize each atom node by its chemical properties, as implemented in the DGL-LifeSci \cite{dgllife} package. Each atom is represented as a 74-dimensional integer vector describing eight pieces of information: the atom type, the atom degree, the number of implicit Hs, formal charge, the number of radical electrons, the atom hybridization, the number of total Hs and whether the atom is aromatic. Similar to the maximum allowed length setting in a protein sequence above, we set a maximum allowed number of nodes $\Theta_d$. Molecules with less nodes will contain virtual nodes with zero padded. As a result, each graph's node feature matrix is denoted as $\mathbf{M}_d \in \mathbb{R}^{\Theta_{d} \times 74}$. Moreover, we use a simple linear transformation to define $\mathbf{X}_d = \mathbf{W}_0\mathbf{M}_d^\top$, leading to a real-valued dense matrix $\mathbf{X}_d \in \mathbb{R}^{\Theta_{d} \times D_d}$ as the input feature.

We employed a three-layer GCN-block to effectively learn the graph representation on drug compounds. GCN generalizes the convolutional operator to an irregular domain. Specifically, we update the atom feature vectors by aggregating their corresponding sets of neighborhood atoms, connected by chemical bonds. This propagation mechanism automatically captures substructure information of a molecule. We keep the node-level drug representation for subsequent explicit learning of local interactions with protein fragments. The drug encoder is written as:

\begin{equation}
    \mathbf{H}_d^{(l+1)} = \sigma(\mathrm{GCN}(\mathbf{\tilde{A}}, \mathbf{W}_{\emph{g}}^{(l)}, \mathbf{b}_{\emph{g}}^{(l)}, \mathbf{H}_p^{(l)})),
\end{equation}
where $\mathbf{W}_{\emph{g}}^{(l)}$ and $\mathbf{b}_{\emph{g}}^{(l)}$ are the GCN's layer-specific learnable weight matrix and bias vector, $\mathbf{\tilde{A}}$ is the adjacency matrix with added self-connections in molecular graph $\mathcal{G}$, and $\mathbf{H}_d^{(l)}$ is the $l$-th hidden node representation with $\mathbf{H}_d^{(0)} = \mathbf{X}_d$.

\textbf{Pairwise interaction learning.} We apply a bilinear attention network module to capture pairwise local interactions between drug and protein. It consists of two layers: (i) A bilinear interaction map to capture pairwise attention weights and (ii) a bilinear pooling layer over the interaction map to extract joint drug-target representation.

Given the third layer's hidden protein and drug representations $\mathbf{H}_p^{(3)} = \{\mathbf{h}^1_p , \mathbf{h}^2_p,...,\mathbf{h}^M_p\}$ and $\mathbf{H}_d^{(3)} = \{\mathbf{h}^1_d , \mathbf{h}^2_d,...,\mathbf{h}^N_d\}$ after separate CNN and GCN encoders, where $M$ and $N$ denote the number of encoded substructures in a protein and atoms in a drug. The bilinear interaction map can obtain a single head pairwise interaction $\mathbf{I} \in \mathbb{R}^{N \times M}$:

\begin{equation}
    \mathbf{I} = ((\mathbf{1} \cdot \mathbf{q}^\top) \circ \sigma((\mathbf{H}_d^{(3)})^\top\mathbf{U})) \cdot \sigma(\mathbf{V}^\top\mathbf{H}_p^{(3)}),
    \label{eq:ba}
\end{equation}
where $\mathbf{U} \in \mathbb{R}^{D_d \times K} $ and $\mathbf{V} \in \mathbb{R}^{D_p \times K}$ are learnable weight matrices for drug and protein representations, $\mathbf{q} \in \mathbb{R}^K$ is a learnable weight vector, $\mathbf{1} \in \mathbb{R}^N$ is a fixed all-ones vector, and $\circ$ denotes Hadamard (element-wise) product. The elements in $\mathbf{I}$ indicate the interaction intensity of respective drug-target sub-structural pairs, with mapping to potential binding sites and molecular substructures. To intuitively understand bilinear interaction, an element $\mathbf{I}_{i,j}$ in Equation (\ref{eq:ba}) can also be written as:

\begin{equation}
    \mathbf{I}_{i,j} = \mathbf{q}^\top(\sigma(\mathbf{U}^\top\mathbf{h}_d^i) \circ \sigma(\mathbf{V}^\top\mathbf{h}_p^j)),
\end{equation}
where $\mathbf{h}_d^i$ is the $i$-th column of $\mathbf{H}_d^{(3)}$ and $\mathbf{h}_p^j$ is the $j$-th column of $\mathbf{H}_p^{(3)}$, respectively denoting the $i$-th and $j$-th sub-structural representations of drug and protein. Therefore, we can see a bilinear interaction as first mapping representations $\mathbf{h}_d^i$ and $\mathbf{h}_p^j$ to a common feature space with weight matrices $\mathbf{U}$ and $\mathbf{V}$, then learn an interaction on Hadamard product and the weight of vector $\mathbf{q}$. In this way, pairwise interactions provide interpretability on the contribution of sub-structural pairs to the predicted result.

To obtain the joint representation $\mathbf{f}^{\prime} \in \mathbb{R}^K$, we introduce a bilinear pooling layer over the interaction map $\mathbf{I}$. Specifically, the $k$-th element of $\mathbf{f}^{\prime}$ is computed as:

\begin{equation}
    \begin{split}
    \mathbf{f}^{\prime}_k &= \sigma((\mathbf{H}_d^{(3)})^\top\mathbf{U})^\top_k \cdot \mathbf{I} \cdot \sigma((\mathbf{H}_p^{(3)})^\top\mathbf{V})_k\\
    &= \sum_{i=1}^N\sum_{j=1}^M \mathbf{I}_{i,j}(\mathbf{h}_d^i)^\top(\mathbf{U}_k\mathbf{V}_k^\top)\mathbf{h}_p^j,
    \end{split}
\end{equation}
where $\mathbf{U}_k$ and $\mathbf{V}_k$ denote the $k$-th column of weight matrices $\mathbf{U}$ and $\mathbf{V}$. Notably, there are no new learnable parameters at this layer. The weight matrices $\mathbf{U}$ and $\mathbf{V}$ are shared with the previous interaction map layer to decrease the number of parameters and alleviate over-fitting. Moreover, we add a sum pooling on the joint representation vector to obtain a compact feature map:

\begin{equation}
    \mathbf{f} = \mathrm{SumPool}(\mathbf{f}^{\prime}, s),
\end{equation}
where the $\rm SumPool(\cdot)$ function is a one-dimensional and non-overlapped sum pooling operation with stride $s$. It reduces the dimensionality of $\mathbf{f}^{\prime} \in \mathbb{R}^{K}$ to $\mathbf{f} \in \mathbb{R}^{K/s}$. Furthermore, we can extend the single pairwise interaction to a multi-head form by calculating multiple bilinear interaction maps. The final joint representation vector is a sum of individual heads. As the weight matrices $\mathbf{U}$ and $\mathbf{V}$ are shared, each additional head only adds one new weight vector $\mathbf{q}$, which is parameter-efficient. In our experiments, the multi-head interaction has a better performance than a single one.

Thus, using the novel bilinear attention mechanism, the model can explicitly learn pairwise local interactions between drug and protein. This interaction module is inspired by and adapted from Kim et al. (2018) \cite{Kim2018BilinearAN} and Yu et al. (2018) \cite{Yu2018BeyondBG}, where two bilinear models are designed for the VQA problem. To compute the interaction probability, we feed the joint representation $\mathbf{f}$ into the decoder, which is one fully connected classification layer followed by a sigmoid function:

\begin{equation}
    p = \mathrm{Sigmoid}(\mathbf{W}_o\mathbf{f} + \mathbf{b}_o),
\end{equation}
where $\mathbf{W}_o$ and $\mathbf{b}_o$ are learnable weight matrix and bias vector. 

Finally, we jointly optimize all learnable parameters by backpropagation. The training objective is to minimize the cross-entropy loss as follows:

\begin{equation}
    \mathcal{L} = -\sum_{i}(y_i\mathrm{log}(p_i) + (1 - y_i)\mathrm{log}(1 - p_i)) + 	\frac{\lambda}{2}\left\| \mathbf{\Theta} \right\|^2_2,
\end{equation}
where $\mathbf{\Theta}$ is the set of all learnable weight matrices and bias vectors above, $y_i$ is the ground-truth label of the $i$-th drug-target pair, $p_i$ is its output probability by the model, and $\lambda$ is a hyperparameter for L2 regularization.

\textbf{Cross-domain adaptation for better generalization.} Machine learning models tend to perform well on similar data from the same distribution (i.e. in-domain), but poorer on dissimilar data with different distribution (i.e. cross-domain). It is a key challenge to improve model performance on cross-domain DTI prediction. In our framework, we embed conditional adversarial domain adaptation (CDAN) to enhance generalization from a source domain with sufficient labeled data to a target domain where only unlabeled data is available. 

Given a source domain $\mathcal{S}_{s} = \{(x_i^{s}, y_i^{s})\}_{i=1}^{N_{s}}$ of $N_{s}$ labeled drug-target pairs and a target domain $\mathcal{S}_{t} = \{x_i^{t}\}_{j=1}^{N_{t}}$ of $N_{t}$ unlabeled drug-target pairs, we leverage CDAN to align their distributions and improve prediction performance across domains. Figure \ref{fig:workflow}c shows the CDAN workflow in our framework, including three key components: the feature extractor $F(\cdot)$, the decoder $G(\cdot)$, and the domain discriminator $D(\cdot)$. We use $F(\cdot)$ to denote the separate feature encoders and bilinear attention network together to generate joint representations of input domain data, i.e., $\mathbf{f}^s_i = F(x^s_i)$ and $\mathbf{f}^t_j = F(x^t_j)$. Next, we use the fully connected classification layer mentioned above followed by a softmax function as $G(\cdot)$ to get a classifier prediction $\mathbf{g}^s_i = G(\mathbf{f}^s_i) \in \mathbb{R}^{2}$ and $\mathbf{g}^t_j = G(\mathbf{f}^t_j) \in \mathbb{R}^{2}$. Furthermore, we apply a multilinear map to embed joint representation $\mathbf{f}$ and classifier prediction $\mathbf{g}$ into a joint conditional representation $\mathbf{h} \in \mathbb{R}^{2K/s}$, which is defined as the flattening of the outer product of the two vectors:

\begin{equation}
    \mathbf{h} = \mathrm{FLATTEN}(\mathbf{f} \otimes \mathbf{g}),
\end{equation}
where $\otimes$ is the outer product.

The multilinear map captures multiplicative interactions  between two independent distributions \cite{Song2009HilbertSE, Song2013RobustLR}. Following the CDAN mechanism, we simultaneously align the joint representation and predicted classification distributions of source and target domains by conditioning the domain discriminator $D(\cdot)$ on the $\mathbf{h}$. The domain discriminator $D(\cdot)$, consisting of a three-layer fully connected networks, learns to distinguish whether a joint conditional representation $\mathbf{h}$ is derived from the source domain or the target domain. On the other hand, the feature extractor $F(\cdot)$ and decoder $G(\cdot)$ are trained to minimize the source domain cross-entropy loss $\mathcal{L}$ with source label information, and simultaneously generate indistinguishable representation $\mathbf{h}$ to confuse the discriminator $D(\cdot)$. As a result, we can formulate the two losses in the cross-domain modeling:

\begin{gather}
    \mathcal{L}_{s}(F, G) = \mathbb{E}_{(x_i^s, y_i^s) \sim \mathcal{S}_{s}} \mathcal{L}(G(F(x_i^s)), y_i^s), \\
    \mathcal{L}_{adv}(F, G, D) = \mathbb{E}_{x_i^t \sim \mathcal{S}_{t}} \mathrm{log}(1 - D(\mathbf{f}_i^t, \mathbf{g}_i^t)) + \mathbb{E}_{x_j^s \sim \mathcal{S}_{s}} log(D(\mathbf{f}_j^s, \mathbf{g}_j^s)),
    \\
\end{gather}
where $\mathcal{L}_{s}$ is the cross-entropy loss on the labeled source domain and $\mathcal{L}_{adv}$ is the adversarial loss for domain discrimination. The optimization problem is written as a minimax paradigm:

\begin{align}
    \begin{split}
    & \rm \mathop{max}\limits_D \mathop{min}\limits_{F, G} \mathcal{L}_{s}(F, G) - \omega\mathcal{L}_{adv}(F, G, D),\\
    \end{split}
\end{align}
where $\omega > 0$ is a hyperparameter to weight $\mathcal{L}_{adv}$. By introducing the adversarial training on $\mathcal{L}_{adv}$, our framework can reduce the data distribution shift between source and target domains, leading to the improved generalization on cross-domain prediction.

\subsection*{Experimental setting}

\textbf{Datasets.} We evaluate DrugBAN and five state-of-the-art baselines on three public DTI datasets: BindingDB, BioSNAP and Human. The BindingDB dataset is a web-accessible database \cite{gilson2016bindingdb} of experimentally validated binding affinities, focusing primarily on the interactions of small drug-like molecules and proteins. We use a low-bias version of the BindingDB dataset constructed in our earlier work Bai et al. (2021) \cite{Bai2021HierarchicalCS}, with the bias-reducing preprocessing steps described in Supplementary Section 2.
The BioSNAP dataset is created from the DrugBank database \cite{Wishart2008DrugBankAK} by Huang et al. (2021) \cite{Huang2021MolTransMI} and Marinka et al. (2018) \cite{biosnapnets}, consisting of 4,510 drugs and 2,181 proteins. It is a balanced dataset with validated positive interactions and an equal number of negative samples randomly obtained from unseen pairs. The Human dataset is constructed by Liu et al. (2015) \cite{Liu2015ImprovingCI}, including highly credible negative samples via an \textit{in silico} screening method. Following previous studies \cite{Chen2020TransformerCPIIC, Tsubaki2019CompoundproteinIP, Zheng2020PredictingDI}, we also use the balanced version of Human dataset containing the same number of positive and negative samples. To mitigate the influence of the hidden data bias \cite{Chen2020TransformerCPIIC}, we use additional cold pair split for performance evaluation on the Human dataset. Supplementary Table 2 shows statistics of the three datasets.

\textbf{Implementation.} DrugBAN is implemented in Python 3.8 and PyTorch 1.7.1 \cite{paszke2017automatic}, along with functions from DGL 0.7.1 \cite{wang2019dgl}, DGLlifeSci 0.2.8 \cite{dgllife}, Scikit-learn 1.0.2 \cite{scikit-learn}, Numpy 1.20.2 \cite{harris2020array}, Pandas 1.2.4 \cite{reback2021pandas} and RDKit 2021.03.2 \cite{rdk}. The batch size is set to be 64 and the Adam optimizer is used with a learning rate of 5e-5. We allow the model to run for at most 100 epochs for all datasets. The best performing model is selected at the epoch giving the best AUROC score on the validation set, which is then used to evaluate the final performance on the test set. The protein feature encoder consists of three 1D-CNN layers with the number of filters [128, 128, 128] and kernel sizes [3, 6, 9]. The drug feature encoder consists of three GCN layers with hidden dimensions [128, 128, 128]. The maximum allowed sequence length for protein is set to be 1200, and the maximum allowed number of atoms for drug molecule is 290. In the bilinear attention module, we only employ two attention heads to provide better interpretability. The latent embedding size $k$ is set to be 768 and the sum pooling window size $s$ is 3. The number of hidden neurons in the fully connected decoder is 512. Our model performance is not sensitive to hyperparameter settings. The configuration details and sensitivity analysis are provided in Supplementary Section 3. We also present a scalability study in Supplementary Section 7.

\textbf{Baselines.} We compare DrugBAN with the following five models on DTI prediction:  (1) Two shallow machine learning methods, support vector machine (SVM) and random forest (RF) applied on the concatenated fingerprint ECFP4 and PSC features; (2) DeepConv-DTI \cite{Lee2019DeepConvDTIPO} that uses CNN and one global max-pooling layer to extract local patterns in protein sequence and a fully connected network to encode drug fingerprint ECFP4; (3) GraphDTA \cite{Nguyen2020GraphDTAPD} that models DTI using graph neural networks to encode drug molecular graph and CNN to encode protein sequence. The learned drug and protein representation vectors are combined with a simple concatenation. To adapt GraphDTA from the original regression task to a binary classification task, we follow the steps in earlier literature \cite{Chen2020TransformerCPIIC, Huang2021MolTransMI} to add a Sigmoid function in its last fully connected layer, and then optimize its parameters with a cross-entropy loss. (4) MolTrans \cite{Huang2021MolTransMI}, a deep learning model adapting transformer architecture to encode drug and protein information, and a CNN-based interactive module to learn sub-structural interaction. For the above deep DTI models, we follow the recommended model hyper-parameter settings described in their original papers.

\section*{Data availability}
The experimental data with each split strategy is available at \url{https://github.com/peizhenbai/DrugBAN/tree/main/datasets}. All data used in this work are from public resource. The BindingDB source is at \url{https://www.bindingdb.org/bind/index.jsp}; The BioSNAP source is at \url{https://github.com/kexinhuang12345/MolTrans} and the Human source is at \url{https://github.com/lifanchen-simm/transformerCPI}.

\section*{Code availability}
The source code and implementation details of DrugBAN are freely available at GitHub repository (\url{https://github.com/peizhenbai/DrugBAN}) and archived on Zenodo (\url{https://doi.org/10.5281/zenodo.7231657}) \cite{peizhen_bai_2022_7415643}.

\section*{Additional information}
\textbf{Competing interests}: the authors declare no competing interests.

\bibliography{main}

\begin{thebibliography}{10}
\urlstyle{rm}
\expandafter\ifx\csname url\endcsname\relax
  \def\url#1{\texttt{#1}}\fi
\expandafter\ifx\csname urlprefix\endcsname\relax\def\urlprefix{URL }\fi
\expandafter\ifx\csname doiprefix\endcsname\relax\def\doiprefix{DOI: }\fi
\providecommand{\bibinfo}[2]{#2}
\providecommand{\eprint}[2][]{\url{#2}}

\bibitem{Luo2017ANI}
\bibinfo{author}{Luo, Y.} \emph{et~al.}
\newblock \bibinfo{journal}{\bibinfo{title}{A network integration approach for
  drug-target interaction prediction and computational drug repositioning from
  heterogeneous information}}.
\newblock {\emph{\JournalTitle{Nature Communications}}}
  \textbf{\bibinfo{volume}{8}} (\bibinfo{year}{2017}).

\bibitem{ztrk2018DeepDTADD}
\bibinfo{author}{{\"O}zt{\"u}rk, H.}, \bibinfo{author}{Olmez, E.~O.} \&
  \bibinfo{author}{{\"O}zg{\"u}r, A.}
\newblock \bibinfo{journal}{\bibinfo{title}{Deep{DTA}: deep drug–target
  binding affinity prediction}}.
\newblock {\emph{\JournalTitle{Bioinformatics}}} \textbf{\bibinfo{volume}{34}},
  \bibinfo{pages}{i821 -- i829} (\bibinfo{year}{2018}).

\bibitem{Yamanishi2008PredictionOD}
\bibinfo{author}{Yamanishi, Y.}, \bibinfo{author}{Araki, M.},
  \bibinfo{author}{Gutteridge, A.}, \bibinfo{author}{Honda, W.} \&
  \bibinfo{author}{Kanehisa, M.}
\newblock \bibinfo{journal}{\bibinfo{title}{Prediction of drug–target
  interaction networks from the integration of chemical and genomic spaces}}.
\newblock {\emph{\JournalTitle{Bioinformatics}}} \textbf{\bibinfo{volume}{24}},
  \bibinfo{pages}{i232 -- i240} (\bibinfo{year}{2008}).

\bibitem{Zitnik2019MachineLF}
\bibinfo{author}{Zitnik, M.} \emph{et~al.}
\newblock \bibinfo{journal}{\bibinfo{title}{Machine learning for integrating
  data in biology and medicine: Principles, {Practice}, and {Opportunities}}}.
\newblock {\emph{\JournalTitle{Information Fusion}}}
  \textbf{\bibinfo{volume}{50}}, \bibinfo{pages}{71--91}
  (\bibinfo{year}{2019}).

\bibitem{Bagherian2021MachineLA}
\bibinfo{author}{Bagherian, M.} \emph{et~al.}
\newblock \bibinfo{journal}{\bibinfo{title}{Machine learning approaches and
  databases for prediction of drug–target interaction: a survey paper}}.
\newblock {\emph{\JournalTitle{Briefings in Bioinformatics}}}
  \textbf{\bibinfo{volume}{22}}, \bibinfo{pages}{247 -- 269}
  (\bibinfo{year}{2021}).

\bibitem{Wen2017DeepLearningBasedDI}
\bibinfo{author}{Wen, M.} \emph{et~al.}
\newblock \bibinfo{journal}{\bibinfo{title}{Deep-learning-based drug-target
  interaction prediction.}}
\newblock {\emph{\JournalTitle{Journal of Proteome Research}}}
  \textbf{\bibinfo{volume}{16 4}}, \bibinfo{pages}{1401--1409}
  (\bibinfo{year}{2017}).

\bibitem{Sieg2019InNO}
\bibinfo{author}{Sieg, J.}, \bibinfo{author}{Flachsenberg, F.} \&
  \bibinfo{author}{Rarey, M.}
\newblock \bibinfo{journal}{\bibinfo{title}{In need of bias control: Evaluating
  chemical data for machine learning in structure-based virtual screening}}.
\newblock {\emph{\JournalTitle{Journal of Chemical Information and Modeling}}}
  \textbf{\bibinfo{volume}{59 3}}, \bibinfo{pages}{947--961}
  (\bibinfo{year}{2019}).

\bibitem{Lim2021ARO}
\bibinfo{author}{Lim, S.} \emph{et~al.}
\newblock \bibinfo{journal}{\bibinfo{title}{A review on compound-protein
  interaction prediction methods: Data, format, representation and model}}.
\newblock {\emph{\JournalTitle{Computational and Structural Biotechnology
  Journal}}} \textbf{\bibinfo{volume}{19}}, \bibinfo{pages}{1541 -- 1556}
  (\bibinfo{year}{2021}).

\bibitem{Gao2018InterpretableDT}
\bibinfo{author}{Gao, K.~Y.} \emph{et~al.}
\newblock \bibinfo{title}{Interpretable drug target prediction using deep
  neural representation}.
\newblock In \emph{\bibinfo{booktitle}{IJCAI}}, \bibinfo{pages}{3371--3377}
  (\bibinfo{year}{2018}).

\bibitem{Bredel2004ChemogenomicsAE}
\bibinfo{author}{Bredel, M.} \& \bibinfo{author}{Jacoby, E.}
\newblock \bibinfo{journal}{\bibinfo{title}{Chemogenomics: an emerging strategy
  for rapid target and drug discovery}}.
\newblock {\emph{\JournalTitle{Nature Reviews Genetics}}}
  \textbf{\bibinfo{volume}{5}}, \bibinfo{pages}{262--275}
  (\bibinfo{year}{2004}).

\bibitem{Lee2019DeepConvDTIPO}
\bibinfo{author}{Lee, I.}, \bibinfo{author}{Keum, J.} \& \bibinfo{author}{Nam,
  H.}
\newblock \bibinfo{journal}{\bibinfo{title}{{DeepConv-DTI}: Prediction of
  drug-target interactions via deep learning with convolution on protein
  sequences}}.
\newblock {\emph{\JournalTitle{PLoS Computational Biology}}}
  \textbf{\bibinfo{volume}{15}} (\bibinfo{year}{2019}).

\bibitem{Hinnerichs2021DTIVoodooML}
\bibinfo{author}{Hinnerichs, T.} \& \bibinfo{author}{Hoehndorf, R.}
\newblock \bibinfo{journal}{\bibinfo{title}{{DTI-Voodoo}: machine learning over
  interaction networks and ontology-based background knowledge predicts
  drug–target interactions}}.
\newblock {\emph{\JournalTitle{Bioinformatics}}} \textbf{\bibinfo{volume}{37}},
  \bibinfo{pages}{4835 -- 4843} (\bibinfo{year}{2021}).

\bibitem{Nguyen2020GraphDTAPD}
\bibinfo{author}{Nguyen, T.} \emph{et~al.}
\newblock \bibinfo{journal}{\bibinfo{title}{Graph{DTA}: Predicting drug-target
  binding affinity with graph neural networks.}}
\newblock {\emph{\JournalTitle{Bioinformatics}}} \textbf{\bibinfo{volume}{37}},
  \bibinfo{pages}{1140--1147} (\bibinfo{year}{2021}).

\bibitem{Tsubaki2019CompoundproteinIP}
\bibinfo{author}{Tsubaki, M.}, \bibinfo{author}{Tomii, K.} \&
  \bibinfo{author}{Sese, J.}
\newblock \bibinfo{journal}{\bibinfo{title}{Compound‐protein interaction
  prediction with end‐to‐end learning of neural networks for graphs and
  sequences}}.
\newblock {\emph{\JournalTitle{Bioinformatics}}} \textbf{\bibinfo{volume}{35}},
  \bibinfo{pages}{309–318} (\bibinfo{year}{2019}).

\bibitem{feng2018padme}
\bibinfo{author}{Feng, Q.}, \bibinfo{author}{Dueva, E.},
  \bibinfo{author}{Cherkasov, A.} \& \bibinfo{author}{Ester, M.}
\newblock \bibinfo{journal}{\bibinfo{title}{P{ADME}: A deep learning-based
  framework for drug-target interaction prediction}}.
\newblock {\emph{\JournalTitle{arXiv preprint arXiv:1807.09741}}}
  (\bibinfo{year}{2018}).

\bibitem{Chen2020TransformerCPIIC}
\bibinfo{author}{Chen, L.} \emph{et~al.}
\newblock \bibinfo{journal}{\bibinfo{title}{{TransformerCPI}: improving
  compound-protein interaction prediction by sequence-based deep learning with
  self-attention mechanism and label reversal experiments}}.
\newblock {\emph{\JournalTitle{Bioinformatics}}}  (\bibinfo{year}{2020}).

\bibitem{Huang2021MolTransMI}
\bibinfo{author}{Huang, K.}, \bibinfo{author}{Xiao, C.},
  \bibinfo{author}{Glass, L.} \& \bibinfo{author}{Sun, J.}
\newblock \bibinfo{journal}{\bibinfo{title}{{MolTrans}: Molecular interaction
  transformer for drug–target interaction prediction}}.
\newblock {\emph{\JournalTitle{Bioinformatics}}} \textbf{\bibinfo{volume}{37}},
  \bibinfo{pages}{830 -- 836} (\bibinfo{year}{2021}).

\bibitem{Schenone2013TargetIA}
\bibinfo{author}{Schenone, M.}, \bibinfo{author}{Danc{\'i}k, V.},
  \bibinfo{author}{Wagner, B.~K.} \& \bibinfo{author}{Clemons, P.~A.}
\newblock \bibinfo{journal}{\bibinfo{title}{Target identification and mechanism
  of action in chemical biology and drug discovery.}}
\newblock {\emph{\JournalTitle{Nature Chemical Biology}}}
  \textbf{\bibinfo{volume}{9 4}}, \bibinfo{pages}{232--40}
  (\bibinfo{year}{2013}).

\bibitem{Ozturk2019WideDTAPO}
\bibinfo{author}{{\"O}zt{\"u}rk, H.}, \bibinfo{author}{Ozkirimli, E.} \&
  \bibinfo{author}{{\"O}zg{\"u}r, A.}
\newblock \bibinfo{journal}{\bibinfo{title}{{WideDTA}: prediction of
  drug-target binding affinity}}.
\newblock {\emph{\JournalTitle{arXiv preprint arXiv:1902.04166}}}
  (\bibinfo{year}{2019}).

\bibitem{Zheng2020PredictingDI}
\bibinfo{author}{Zheng, S.}, \bibinfo{author}{Li, Y.}, \bibinfo{author}{Chen,
  S.}, \bibinfo{author}{Xu, J.} \& \bibinfo{author}{Yang, Y.}
\newblock \bibinfo{journal}{\bibinfo{title}{Predicting drug–protein
  interaction using quasi-visual question answering system}}.
\newblock {\emph{\JournalTitle{Nature Machine Intelligence}}}
  \textbf{\bibinfo{volume}{2}}, \bibinfo{pages}{134--140}
  (\bibinfo{year}{2020}).

\bibitem{Long2018ConditionalAD}
\bibinfo{author}{Long, M.}, \bibinfo{author}{Cao, Z.}, \bibinfo{author}{Wang,
  J.} \& \bibinfo{author}{Jordan, M.~I.}
\newblock \bibinfo{title}{Conditional {Adversarial} {Domain} {Adaptation}}.
\newblock In \emph{\bibinfo{booktitle}{NeurIPS}} (\bibinfo{year}{2018}).

\bibitem{Kim2017}
\bibinfo{author}{Kim, J.-H.} \emph{et~al.}
\newblock \bibinfo{title}{{Hadamard Product for Low-rank Bilinear Pooling}}.
\newblock In \emph{\bibinfo{booktitle}{ICLR}} (\bibinfo{year}{2017}).

\bibitem{Abbasi2020DeepCDADC}
\bibinfo{author}{Abbasi, K.} \emph{et~al.}
\newblock \bibinfo{journal}{\bibinfo{title}{{DeepCDA}: deep cross-domain
  compound--protein affinity prediction through lstm and convolutional neural
  networks}}.
\newblock {\emph{\JournalTitle{Bioinformatics}}} \textbf{\bibinfo{volume}{36}},
  \bibinfo{pages}{4633--4642} (\bibinfo{year}{2020}).

\bibitem{kao2021toward}
\bibinfo{author}{Kao, P.-Y.}, \bibinfo{author}{Kao, S.-M.},
  \bibinfo{author}{Huang, N.-L.} \& \bibinfo{author}{Lin, Y.-C.}
\newblock \bibinfo{title}{Toward drug-target interaction prediction via
  ensemble modeling and transfer learning}.
\newblock In \emph{\bibinfo{booktitle}{2021 IEEE International Conference on
  Bioinformatics and Biomedicine (BIBM)}}, \bibinfo{pages}{2384--2391}
  (\bibinfo{organization}{IEEE}, \bibinfo{year}{2021}).

\bibitem{abbasi2021deep}
\bibinfo{author}{Abbasi, K.}, \bibinfo{author}{Razzaghi, P.},
  \bibinfo{author}{Poso, A.}, \bibinfo{author}{Ghanbari-Ara, S.} \&
  \bibinfo{author}{Masoudi-Nejad, A.}
\newblock \bibinfo{journal}{\bibinfo{title}{Deep learning in drug target
  interaction prediction: current and future perspectives}}.
\newblock {\emph{\JournalTitle{Current Medicinal Chemistry}}}
  \textbf{\bibinfo{volume}{28}}, \bibinfo{pages}{2100--2113}
  (\bibinfo{year}{2021}).

\bibitem{Kipf2017SemiSupervisedCW}
\bibinfo{author}{Kipf, T.} \& \bibinfo{author}{Welling, M.}
\newblock \bibinfo{title}{Semi-supervised classification with graph
  convolutional networks}.
\newblock In \emph{\bibinfo{booktitle}{ICLR}} (\bibinfo{year}{2017}).

\bibitem{Yu2018BeyondBG}
\bibinfo{author}{Yu, Z.}, \bibinfo{author}{Yu, J.}, \bibinfo{author}{Xiang,
  C.}, \bibinfo{author}{Fan, J.} \& \bibinfo{author}{Tao, D.}
\newblock \bibinfo{journal}{\bibinfo{title}{Beyond bilinear: Generalized
  multimodal factorized high-order pooling for visual question answering}}.
\newblock {\emph{\JournalTitle{IEEE Trans. Neural Netw. Learn. Syst.}}}
  \textbf{\bibinfo{volume}{29}}, \bibinfo{pages}{5947--5959}
  (\bibinfo{year}{2018}).

\bibitem{Kim2018BilinearAN}
\bibinfo{author}{Kim, J.-H.}, \bibinfo{author}{Jun, J.} \&
  \bibinfo{author}{Zhang, B.-T.}
\newblock \bibinfo{title}{Bilinear {Attention} {Networks}}.
\newblock In \emph{\bibinfo{booktitle}{NeurIPS}} (\bibinfo{year}{2018}).

\bibitem{Weininger1988SMILESAC}
\bibinfo{author}{Weininger, D.}
\newblock \bibinfo{journal}{\bibinfo{title}{{SMILES}, a chemical language and
  information system. 1. introduction to methodology and encoding rules}}.
\newblock {\emph{\JournalTitle{Journal of Chemical Information and Computer
  Sciences}}} \textbf{\bibinfo{volume}{28}}, \bibinfo{pages}{31--36}
  (\bibinfo{year}{1988}).

\bibitem{Liu2007BindingDBAW}
\bibinfo{author}{Liu, T.}, \bibinfo{author}{Lin, Y.}, \bibinfo{author}{Wen,
  X.}, \bibinfo{author}{Jorissen, R.~N.} \& \bibinfo{author}{Gilson, M.~K.}
\newblock \bibinfo{journal}{\bibinfo{title}{Bindingdb: a web-accessible
  database of experimentally determined protein–ligand binding affinities}}.
\newblock {\emph{\JournalTitle{Nucleic Acids Research}}}
  \textbf{\bibinfo{volume}{35}}, \bibinfo{pages}{D198 -- D201}
  (\bibinfo{year}{2007}).

\bibitem{biosnapnets}
\bibinfo{author}{Zitnik, M.}, \bibinfo{author}{Sosi\v{c}, R.},
  \bibinfo{author}{Maheshwari, S.} \& \bibinfo{author}{Leskovec, J.}
\newblock \bibinfo{title}{{BioSNAP Datasets}: {Stanford} biomedical network
  dataset collection} (\bibinfo{year}{2018}).

\bibitem{Liu2015ImprovingCI}
\bibinfo{author}{Liu, H.}, \bibinfo{author}{Sun, J.}, \bibinfo{author}{Guan,
  J.}, \bibinfo{author}{Zheng, J.} \& \bibinfo{author}{Zhou, S.}
\newblock \bibinfo{journal}{\bibinfo{title}{Improving compound–protein
  interaction prediction by building up highly credible negative samples}}.
\newblock {\emph{\JournalTitle{Bioinformatics}}} \textbf{\bibinfo{volume}{31}},
  \bibinfo{pages}{i221 -- i229} (\bibinfo{year}{2015}).

\bibitem{Rogers2010ExtendedConnectivityF}
\bibinfo{author}{Rogers, D.} \& \bibinfo{author}{Hahn, M.}
\newblock \bibinfo{journal}{\bibinfo{title}{Extended-connectivity
  fingerprints}}.
\newblock {\emph{\JournalTitle{Journal of Chemical Information and Modeling}}}
  \textbf{\bibinfo{volume}{50 5}}, \bibinfo{pages}{742--54}
  (\bibinfo{year}{2010}).

\bibitem{Cao2013propyAT}
\bibinfo{author}{Cao, D.}, \bibinfo{author}{Xu, Q.} \& \bibinfo{author}{Liang,
  Y.}
\newblock \bibinfo{journal}{\bibinfo{title}{Propy: a tool to generate various
  modes of chou's pseaac}}.
\newblock {\emph{\JournalTitle{Bioinformatics}}} \textbf{\bibinfo{volume}{29
  7}}, \bibinfo{pages}{960--2} (\bibinfo{year}{2013}).

\bibitem{cortes1995support}
\bibinfo{author}{Cortes, C.} \& \bibinfo{author}{Vapnik, V.}
\newblock \bibinfo{journal}{\bibinfo{title}{Support-vector networks}}.
\newblock {\emph{\JournalTitle{Machine learning}}}
  \textbf{\bibinfo{volume}{20}}, \bibinfo{pages}{273--297}
  (\bibinfo{year}{1995}).

\bibitem{ho1995random}
\bibinfo{author}{Ho, T.~K.}
\newblock \bibinfo{title}{Random decision forests}.
\newblock In \emph{\bibinfo{booktitle}{Proceedings of 3rd International
  Conference on Document Analysis and Recognition}}, vol.~\bibinfo{volume}{1},
  \bibinfo{pages}{278--282} (\bibinfo{year}{1995}).

\bibitem{Ganin2016DomainAdversarialTO}
\bibinfo{author}{Ganin, Y.} \emph{et~al.}
\newblock \bibinfo{title}{Domain-adversarial training of neural networks}.
\newblock In \emph{\bibinfo{booktitle}{J. Mach. Learn. Res.}}
  (\bibinfo{year}{2016}).

\bibitem{rdk}
\bibinfo{author}{{Greg Landrum et al}}.
\newblock \bibinfo{title}{{RDKit: Open-source cheminformatics}}
  (\bibinfo{year}{2006}).

\bibitem{moe}
\bibinfo{author}{{Molecular Operating Environment (MOE)}}.
\newblock \bibinfo{title}{{2020.09 Chemical Computing Group ULC, 1010 Sherbooke
  St. West, Suite \#910, Montreal, QC, Canada, H3A 2R7}}
  (\bibinfo{year}{2022}).

\bibitem{kazokaite2019engineered}
\bibinfo{author}{Kazokait{\.e}, J.} \emph{et~al.}
\newblock \bibinfo{journal}{\bibinfo{title}{Engineered carbonic anhydrase
  vi-mimic enzyme switched the structure and affinities of inhibitors}}.
\newblock {\emph{\JournalTitle{Scientific reports}}}
  \textbf{\bibinfo{volume}{9}}, \bibinfo{pages}{1--17} (\bibinfo{year}{2019}).

\bibitem{rai2017discovery}
\bibinfo{author}{Rai, G.} \emph{et~al.}
\newblock \bibinfo{journal}{\bibinfo{title}{Discovery and optimization of
  potent, cell-active pyrazole-based inhibitors of lactate dehydrogenase
  (ldh)}}.
\newblock {\emph{\JournalTitle{Journal of medicinal chemistry}}}
  \textbf{\bibinfo{volume}{60}}, \bibinfo{pages}{9184--9204}
  (\bibinfo{year}{2017}).

\bibitem{fenalti2014molecular}
\bibinfo{author}{Fenalti, G.} \emph{et~al.}
\newblock \bibinfo{journal}{\bibinfo{title}{Molecular control of
  $\delta$-opioid receptor signalling}}.
\newblock {\emph{\JournalTitle{Nature}}} \textbf{\bibinfo{volume}{506}},
  \bibinfo{pages}{191--196} (\bibinfo{year}{2014}).

\bibitem{berman2000protein}
\bibinfo{author}{Berman, H.~M.} \emph{et~al.}
\newblock \bibinfo{journal}{\bibinfo{title}{The protein data bank}}.
\newblock {\emph{\JournalTitle{Nucleic acids research}}}
  \textbf{\bibinfo{volume}{28}}, \bibinfo{pages}{235--242}
  (\bibinfo{year}{2000}).

\bibitem{Jumper2021HighlyAP}
\bibinfo{author}{Jumper, J.~M.} \emph{et~al.}
\newblock \bibinfo{journal}{\bibinfo{title}{Highly accurate protein structure
  prediction with alphafold}}.
\newblock {\emph{\JournalTitle{Nature}}} \textbf{\bibinfo{volume}{596}},
  \bibinfo{pages}{583 -- 589} (\bibinfo{year}{2021}).

\bibitem{Pan2010ASO}
\bibinfo{author}{Pan, S.~J.} \& \bibinfo{author}{Yang, Q.}
\newblock \bibinfo{journal}{\bibinfo{title}{A survey on transfer learning}}.
\newblock {\emph{\JournalTitle{IEEE Trans. Knowl. Data Eng.}}}
  \textbf{\bibinfo{volume}{22}}, \bibinfo{pages}{1345--1359}
  (\bibinfo{year}{2010}).

\bibitem{Gong2013ConnectingTD}
\bibinfo{author}{Gong, B.}, \bibinfo{author}{Grauman, K.} \&
  \bibinfo{author}{Sha, F.}
\newblock \bibinfo{title}{Connecting the dots with landmarks: Discriminatively
  learning domain-invariant features for unsupervised domain adaptation}.
\newblock In \emph{\bibinfo{booktitle}{ICML}} (\bibinfo{year}{2013}).

\bibitem{Huang2006CorrectingSS}
\bibinfo{author}{Huang, J.}, \bibinfo{author}{Smola, A.},
  \bibinfo{author}{Gretton, A.}, \bibinfo{author}{Borgwardt, K.~M.} \&
  \bibinfo{author}{Sch{\"o}lkopf, B.}
\newblock \bibinfo{title}{Correcting sample selection bias by unlabeled data}.
\newblock In \emph{\bibinfo{booktitle}{NIPS}} (\bibinfo{year}{2006}).

\bibitem{dgllife}
\bibinfo{author}{Li, M.} \emph{et~al.}
\newblock \bibinfo{journal}{\bibinfo{title}{{DGL-LifeSci: An Open-Source
  Toolkit for Deep Learning on Graphs in Life Science}}}.
\newblock {\emph{\JournalTitle{ACS Omega}}}  (\bibinfo{year}{2021}).

\bibitem{Song2009HilbertSE}
\bibinfo{author}{Song, L.}, \bibinfo{author}{Huang, J.},
  \bibinfo{author}{Smola, A.} \& \bibinfo{author}{Fukumizu, K.}
\newblock \bibinfo{title}{Hilbert space embeddings of conditional distributions
  with applications to dynamical systems}.
\newblock In \emph{\bibinfo{booktitle}{ICML}} (\bibinfo{year}{2009}).

\bibitem{Song2013RobustLR}
\bibinfo{author}{Song, L.} \& \bibinfo{author}{Dai, B.}
\newblock \bibinfo{title}{Robust low rank kernel embeddings of multivariate
  distributions}.
\newblock In \emph{\bibinfo{booktitle}{NIPS}} (\bibinfo{year}{2013}).

\bibitem{gilson2016bindingdb}
\bibinfo{author}{Gilson, M.~K.} \emph{et~al.}
\newblock \bibinfo{journal}{\bibinfo{title}{Binding{DB} in 2015: a public
  database for medicinal chemistry, computational chemistry and systems
  pharmacology}}.
\newblock {\emph{\JournalTitle{Nucleic acids research}}}
  \textbf{\bibinfo{volume}{44}}, \bibinfo{pages}{D1045--D1053}
  (\bibinfo{year}{2016}).

\bibitem{Bai2021HierarchicalCS}
\bibinfo{author}{Bai, P.} \emph{et~al.}
\newblock \bibinfo{journal}{\bibinfo{title}{Hierarchical clustering split for
  low-bias evaluation of drug-target interaction prediction}}.
\newblock {\emph{\JournalTitle{2021 IEEE International Conference on
  Bioinformatics and Biomedicine (BIBM)}}} \bibinfo{pages}{641--644}
  (\bibinfo{year}{2021}).

\bibitem{Wishart2008DrugBankAK}
\bibinfo{author}{Wishart, D.~S.} \emph{et~al.}
\newblock \bibinfo{journal}{\bibinfo{title}{Drug{Bank}: a knowledgebase for
  drugs, drug actions and drug targets}}.
\newblock {\emph{\JournalTitle{Nucleic Acids Research}}}
  \textbf{\bibinfo{volume}{36}}, \bibinfo{pages}{D901 -- D906}
  (\bibinfo{year}{2008}).

\bibitem{paszke2017automatic}
\bibinfo{author}{Paszke, A.} \emph{et~al.}
\newblock \bibinfo{title}{Automatic differentiation in {PyTorch}}
  (\bibinfo{year}{2017}).

\bibitem{wang2019dgl}
\bibinfo{author}{Wang, M.} \emph{et~al.}
\newblock \bibinfo{journal}{\bibinfo{title}{{Deep Graph Library: A
  Graph-Centric, Highly-Performant Package for Graph Neural Networks}}}.
\newblock {\emph{\JournalTitle{arXiv preprint arXiv:1909.01315}}}
  (\bibinfo{year}{2019}).

\bibitem{scikit-learn}
\bibinfo{author}{Pedregosa, F.} \emph{et~al.}
\newblock \bibinfo{journal}{\bibinfo{title}{{Scikit-learn: Machine Learning in
  {P}ython}}}.
\newblock {\emph{\JournalTitle{Journal of Machine Learning Research}}}
  \textbf{\bibinfo{volume}{12}}, \bibinfo{pages}{2825--2830}
  (\bibinfo{year}{2011}).

\bibitem{harris2020array}
\bibinfo{author}{Harris, C.~R.} \emph{et~al.}
\newblock \bibinfo{journal}{\bibinfo{title}{{Array programming with {NumPy}}}}.
\newblock {\emph{\JournalTitle{Nature}}} \textbf{\bibinfo{volume}{585}},
  \bibinfo{pages}{357--362}, \doiprefix\url{10.1038/s41586-020-2649-2}
  (\bibinfo{year}{2020}).

\bibitem{reback2021pandas}
\bibinfo{author}{{The pandas development team}}.
\newblock \bibinfo{title}{pandas-dev/pandas: Pandas 1.2.4},
  \doiprefix\url{10.5281/zenodo.4681666} (\bibinfo{year}{2021}).

\bibitem{peizhen_bai_2022_7415643}
\bibinfo{author}{Bai, P.} \& \bibinfo{author}{Lu, H.}
\newblock \bibinfo{title}{peizhenbai/drugban: v1.2.0},
  \doiprefix\url{10.5281/zenodo.7415643} (\bibinfo{year}{2022}).

\bibitem{wang2016improving}
\bibinfo{author}{Wang, Z.}, \bibinfo{author}{Liang, L.}, \bibinfo{author}{Yin,
  Z.} \& \bibinfo{author}{Lin, J.}
\newblock \bibinfo{journal}{\bibinfo{title}{Improving chemical similarity
  ensemble approach in target prediction}}.
\newblock {\emph{\JournalTitle{Journal of cheminformatics}}}
  \textbf{\bibinfo{volume}{8}}, \bibinfo{pages}{1--10} (\bibinfo{year}{2016}).

\end{thebibliography}

\newpage
\section*{Supplementary Material}

\subsection*{S1. Clustering-based pair split strategy}
\label{sec:cluster}
As mentioned in the main text, we separately cluster drug compounds and target proteins of the BindingDB and BioSNAP datasets for cross-domain performance evaluation. Specifically,  we choose the single-linkage clustering, a bottom-up hierarchical clustering to ensure that the distances between samples in different clusters are always larger than a pre-defined distance, i.e., minimum distance threshold $\gamma$. This property can prevent clusters from being too close to help to generate the cross-domain scenario.

We use binarized ECFP4 feature to represent drug compounds, and integral PSC feature to represent target proteins. For accurately measuring the pairwise distance, we use the Jaccard distance and cosine distance on ECFP4 and PSC, respectively. We choose $\gamma = 0.5$ in both drug and protein clusterings since this choice can prevent over-large clusters and be ensure separate dissimilar samples. We obtained 2,780 clusters of drugs and 1,693 clusters of proteins for the BindingDB dataset, and 2,387 clusters of drugs and 1,978 clusters of proteins for the BioSNAP dataset. Table \ref{ten_cluster} shows the number of samples in the ten largest clusters of the clustering results. It shows that BindingDB has a more balanced cluster distribution than BioSNAP in drug clustering. In addition, the protein clustering result tends to generate many small clusters with only a few proteins in both datasets, indicating that the average similarity between proteins is lower than that between drugs. We randomly select 60\% drug clusters and 60\% protein clusters from clustering result, and regard all associated drug-target pairs with them as source domain data. The associated pairs in the remaining clusters are considered to be source domain data. We conduct five independent clustering-based pair splits with different random seeds for downstream model training and evaluation. Clustering-based pair split allows quantitatively constructing cross-domain tasks by considering the similarity between drugs or proteins.
\begin{table*}[hbt!]
\centering
\caption{\centering Size of the ten largest clusters in the BindingDB and BioSNAP datasets generated by the clustering-based pair split.}
 \setlength{\tabcolsep}{3mm}{\begin{tabular}{llllllllllll}
\toprule
Dataset & Object & \# 1 & \# 2 & \# 3 & \# 4 & \# 5 & \# 6 & \# 7 & \# 8 & \# 9 & \# 10\\ \midrule
BindingDB & Drug & 598 & 460 & 304 & 290 & 253 & 250 & 203 & 202 & 198 & 158 \\
BioSNAP & Drug & 294 & 267 & 75 & 68 & 36 & 35 & 28 & 26 & 24 & 24 \\
BindingDB & Protein & 17 & 15 & 15 & 12 & 10 & 10 & 10 & 9 & 9 & 8 \\
BioSNAP & Protein & 8 & 8 & 8 & 6 & 5 & 4 & 4 & 4 & 4 & 4 \\
\bottomrule
\end{tabular}}
\label{ten_cluster}
\end{table*}

\subsection*{S2. Dataset statistics, notations, and preprocessing steps}
\label{sec:datanota}

Table \ref{dataset} shows the statistics of experimental datasets and Table \ref{Notation} lists the notations used in this paper with descriptions. The BioSNAP and Human datasets were created by Huang et al. (2021)\cite{Huang2021MolTransMI} and Liu et al. (2015) \cite{Liu2015ImprovingCI}, respectively. For the BindingDB dataset, we created a low-bias version from the BindingDB database source \cite{gilson2016bindingdb} following the bias-reducing preprocessing steps in our earlier work \cite{Bai2021HierarchicalCS}: i) We considered a drug-target pair to be positive only if its IC50 is less than 100 nM, and negative only if its IC50 was greater than 10,000 nM, giving a 100-fold difference to reduce class label noise. These IC50 thresholds were selected following earlier works \cite{Gao2018InterpretableDT, wang2016improving}. ii) We removed all DTI pairs where the drugs only had one type of pairs (positive or negative) to improve drug-wise pair class balance and reduce hidden ligand bias that can lead to the correct predictions based only on drug features.

\begin{table}[hbt!]
\centering
\caption{\centering Experimental dataset statistics}
\begin{tabular}{lrrrr}
\toprule
Dataset & \# Drugs & \# Proteins & \# Interactions   \\ \midrule
BindingDB \cite{Bai2021HierarchicalCS}  & 14,643 & 2,623 & 49,199   \\
BioSNAP \cite{Huang2021MolTransMI} & 4,510 & 2,181 & 27,464\\
Human \cite{Liu2015ImprovingCI} & 2,726 & 2,001 & 6,728\\ 
\bottomrule
\end{tabular}
\label{dataset}
\end{table}

\begin{table*}[ht]
\renewcommand\arraystretch{1.2}
\centering
\caption{\centering Notations and descriptions}
 \setlength{\tabcolsep}{1mm}{\begin{tabular}{ll}
\toprule
Notations & Description \\ \midrule
$\mathbf{E}_{p} \in \mathbb{R}^{23 \times D_p}$ & protein amino acid embedding matrix \\
$\mathbf{f} \in \mathbb{R}^{K/s}$ &  drug-target joint representation\\
$F(\cdot), G(\cdot), D(\cdot)$ & feature extractor, decoder and domain discriminator in CDAN\\
$\mathbf{g} \in \mathbb{R}^{2}$ & output interaction probability by softmax function\\
$\mathbf{H}_p^{(l)}$, $\mathbf{H}_d^{(l)}$ & hidden representation for protein (drug) in $l$-th CNN (GCN) layer\\
$\mathbf{I} \in \mathbb{R}^{N \times M}$ & pair-wise interaction matrix between drug and protein substructures\\
$\mathbf{M}_d \in \mathbb{R}^{\Theta_{d} \times 74}$ & drug node feature matrix by its chemical properties\\
$p \in \mathbb{R}^{1}$ & output interaction probability by Sigmoid function\\
$\mathcal{P}$, $\mathcal{G}$ & protein amino acid sequence, drug 2D molecular graph \\
$\mathbf{q} \in \mathbb{R}^K$ &  weight vector for bilinear transformation\\
$\mathbf{U} \in \mathbb{R}^{D_d \times K}$ & the weight matrix for encoded drug representation\\
$\mathbf{V} \in \mathbb{R}^{D_p \times K}$ & the weight matrix for encoded protein representation \\
$\mathbf{W}_c$, $\mathbf{b}_c$ & the weight matrix and bias for protein CNN encoder\\
$\mathbf{W}_g$, $\mathbf{b}_g$ & the weight matrix and bias for drug GCN encoder\\
$\mathbf{W}_o$, $\mathbf{b}_o$ & the weight matrix and bias for decoder\\
$\mathbf{X}_p \in \mathbb{R}^{\Theta_{p} \times D_p}$ & latent protein matrix representation \\
$\mathbf{X}_d \in \mathbb{R}^{\Theta_{d} \times D_d}$ & latent drug matrix representation \\
\bottomrule
\end{tabular}}
\label{Notation}
\end{table*}

\subsection*{S3. Hyperparameter setting and sensitivity analysis}
\label{sec:hyper}

\begin{figure}[hbt!]
\centering
    \includegraphics[width=1\textwidth]{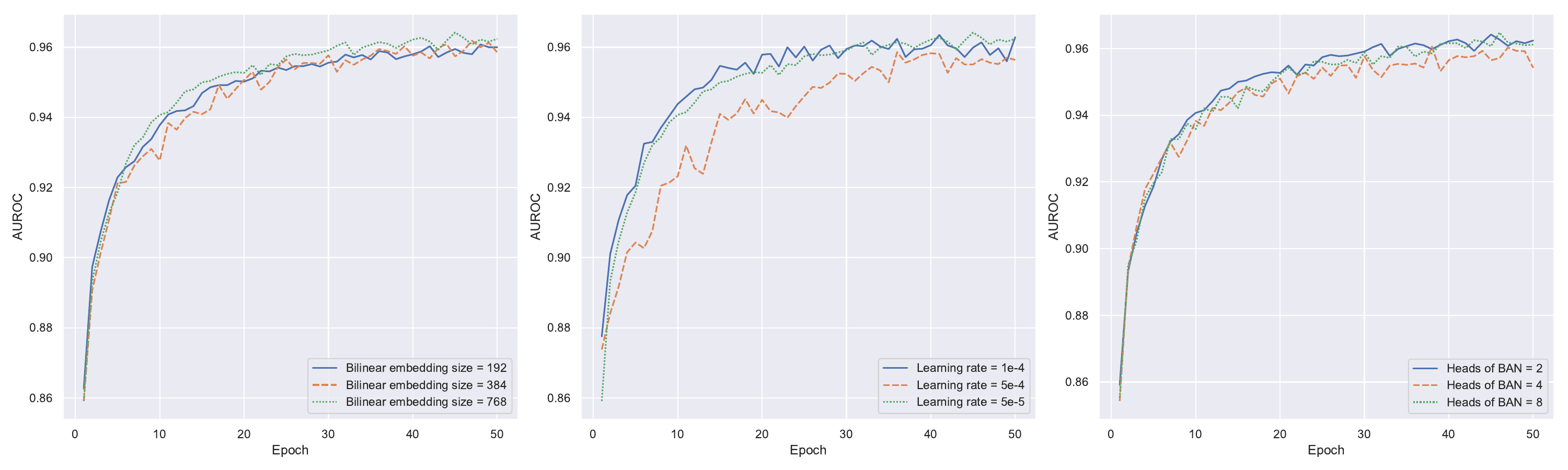}
    \caption{Learning curves with the different choices of hyperparameters on the BindingDB validation set.}
    \label{fig:sen}
\end{figure}

Table \ref{Hyper} shows a list of model hyperparameters and their values used in experiment. As our model performance is not sensitive to hyperparameter setting, we use the same hyperparamters on all experimental datasets (BindingDB, BioSNAP and Human). Figure \ref{fig:sen} illustrates the learning curves with the different choices of hyperparameters on the BindingDB validation set, including bilinear embedding size, learning rate and heads of attention. It shows that the performance differences are not large and typically converges between 30 and 40 epochs. 

\begin{table*}[hbt!]
\centering
\caption{\centering DrugBAN hyperparameter configuration}
 \setlength{\tabcolsep}{5mm}{\begin{tabular}{lll}
\toprule
Module & Hyperparameter & Value \\ \midrule
Optimizer & Learning rate & 5e-5 \\
Mini-batch          & Batch size & 64 \\
Three-layer CNN protein encoder & Initial amino acid embedding & 128 \\
                    & Number of filters & [128, 128, 128] \\
                    & Kernel size & [3, 6, 9] \\
Three-layer GCN drug encoder & Initial atom embedding & 128 \\
                 & Hidden node dimensions  & [128, 128, 128] \\
Bilinear interaction attention & Heads of bilinear attention & 2 \\
                               & Bilinear embedding size & 768 \\
                               & Sum pooling window size & 3 \\
Fully connected decoder & Number of hidden neurons & 512 \\
Discriminator & Number of hidden neurons & 256 \\
\bottomrule
\end{tabular}}
\label{Hyper}
\end{table*}

\subsection*{S4. Performance comparison across different protein families}
\label{sec:pro_families}

\begin{table}[hbt!]
\centering
{\caption{\centering Number of interactions for major protein families in the test sets}
\begin{tabular}{lrrrr}
\toprule
Dataset & \# Enzymes & \# GPCRs & \# Ion channels & \# NHRs \\ \midrule
BindingDB  & 5,277 & 472 & 440  & 144   \\
BioSNAP & 1,956 & 536 & 510  & 103  \\
\bottomrule
\end{tabular}
\label{family_data}}
\end{table}

\begin{figure}[hbt!]
\centering
% \hspace{-2cm}  
    \includegraphics[width=0.8\textwidth]{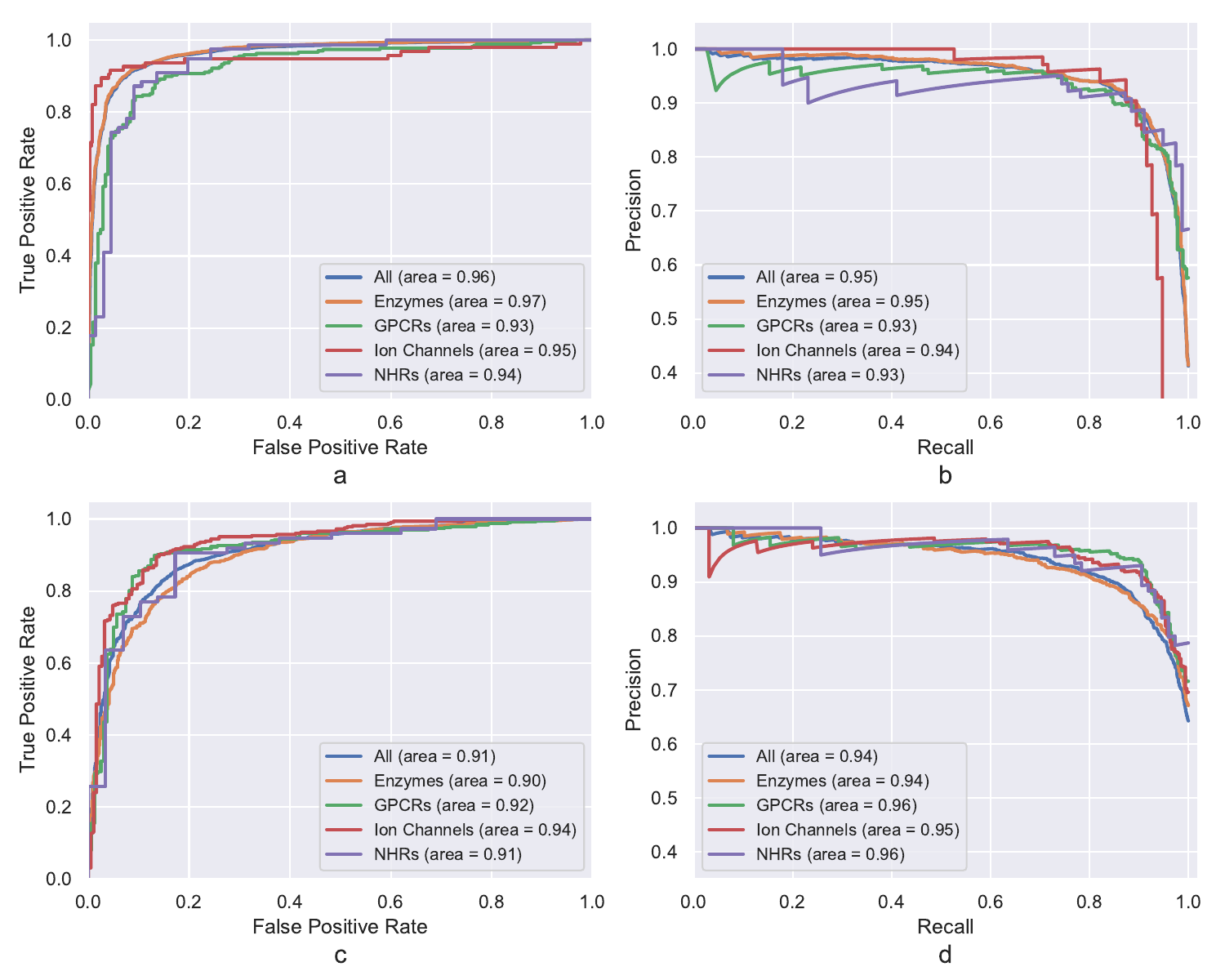}
    \caption{\textbf{DrugBAN performance on different protein families.} \textbf{(a)} AUROC curves on the BindingDB dataset. \textbf{(b)} AUPRC curves on the BindingDB dataset. \textbf{(c)} AUROC curves on the BioSNAP dataset. \textbf{(d)} AUPRC curves on the BioSNAP dataset.}
    \label{fig:protein_families}
\end{figure}

We conduct experiments to study the performance of DrugBAN on different protein families. Following the previous studies  \cite{Huang2021MolTransMI, Yamanishi2008PredictionOD}, we select four major protein families: enzymes, G protein-coupled receptors (GPCRs), ion channels  and nuclear hormone receptors (NHRs). We randomly retrieve one in-domain test set of BindingDB and BioSNAP respectively, and map their proteins to the four protein families using GtoPdb database (\url{https://www.guidetopharmacology.org/targets.jsp}). Table \ref{family_data} presents the number of interactions for each protein family in the test sets. Figure \ref{fig:protein_families} shows the performance (AUROC and AUPRC) varying only slightly given different protein families.

\subsection*{S5 Performance comparison on unseen drugs/targets}
\label{sec:unseen_cases}

\begin{table*}[hbt!]
\centering
{\small
\caption{\centering Performance (average AUROC over five random runs) comparison on the BindingDB and BioSNAP datasets with random split, unseen drug, and unseen target settings (\textbf{Best}, \underline{Second Best}).}
\setlength{\tabcolsep}{5mm}{\begin{tabular}{l|cccc}
\toprule
Setting                & DeepConv-DTI \cite{Lee2019DeepConvDTIPO} & GraphDTA \cite{Nguyen2020GraphDTAPD} & MolTrans \cite{Huang2021MolTransMI} & DrugBAN \\ \midrule
              & \multicolumn{4}{c}{BindingDB} \\
Random Split   & 0.945$\pm$0.002 & 0.951$\pm$0.002 &\underline{0.952$\pm$0.002} & \textbf{0.960$\pm$0.001} \\
Unseen Drug   & 0.943$\pm$0.004 & \underline{0.950$\pm$0.004} & 0.945$\pm$0.004 & \textbf{0.959$\pm$0.002} \\
Unseen Target & 0.627$\pm$0.070 & \underline{0.670$\pm$0.023} & 0.661$\pm$0.037 & \textbf{0.692$\pm$0.038} \\
& \multicolumn{4}{c}{BioSNAP} \\
Random Split   & 0.886$\pm$0.006 & 0.887$\pm$0.008 & \underline{0.895$\pm$0.004} & \textbf{0.903$\pm$0.005} \\
Unseen Drug   & 0.856$\pm$0.005 & \underline{0.858$\pm$0.007} & 0.856$\pm$0.008 & \textbf{0.886$\pm$0.005} \\
Unseen Target & 0.692$\pm$0.017 & 0.704$\pm$0.010 & \textbf{0.714$\pm$0.014} & \underline{0.710$\pm$0.016}  \\
\bottomrule
\end{tabular}}
\label{unseen}}
\end{table*}

To study how DrugBAN and other deep learning baselines perform on unseen drugs/targets, we conduct additional experiments on BindingDB and BioSNAP. For each dataset, we randomly select 20\% drugs/target proteins. Then we evaluate predictive performance on all DTI pairs associated with these drugs/target proteins (70\% as test set for evaluation and 30\% as validation set for determining early stopping), and the rest pairs as training set for model optimization. Each unseen setting has five independent runs. Table \ref{unseen} presents the AUROC results on the test sets, including the results on the usual random split for comparison. DrugBAN achieves the best performance in five of the six settings, while its performance in the unseen target setting of BioSNAP is also very competitive.

We need to point out that the model performance under the unseen drug setting only dropped slightly compared to that under the random split for all methods on BindingDB. This is because there are many highly similar molecules in the DTI datasets, and naive unseen drug setting does not distinguish them. A better strategy is the clustering-based split strategy in our previous study to alleviate this issue, leading to a more challenging cross-domain task.

\subsection*{S6 Performance comparison with high fraction of missing data}
\label{sec:missing_data}

\begin{table*}[hbt!]
\centering
{\small
\caption{\centering Performance comparison (average AUROC over five random runs) on the BindingDB and BioSNAP datasets with high fraction of missing data (\textbf{Best}, \underline{Second Best})}
\setlength{\tabcolsep}{5mm}{\begin{tabular}{c|cccc}
\toprule
Missing (\%)                & DeepConv-DTI \cite{Lee2019DeepConvDTIPO} & GraphDTA \cite{Nguyen2020GraphDTAPD} & MolTrans \cite{Huang2021MolTransMI} & DrugBAN \\ \midrule
              & \multicolumn{4}{c}{BindingDB} \\
95   & 0.773$\pm$0.005 & 0.831$\pm$0.002 & \underline{0.846$\pm$0.004} & \textbf{0.856$\pm$0.003}\\              
90   & 0.840$\pm$0.002 & 0.867$\pm$0.002 & \underline{0.874$\pm$0.003} & \textbf{0.887$\pm$0.004}\\
80   & 0.877$\pm$0.002 & 0.897$\pm$0.003 & \underline{0.905$\pm$0.001} & \textbf{0.920$\pm$0.003}\\
70   & 0.890$\pm$0.005 & 0.916$\pm$0.002 & \underline{0.923$\pm$0.001} & \textbf{0.934$\pm$0.001}\\
& \multicolumn{4}{c}{BioSNAP} \\
95   & 0.710$\pm$0.005 & \underline{0.768$\pm$0.005} & 0.767$\pm$0.006 & \textbf{0.770$\pm$0.008}\\
90   & 0.781$\pm$0.003 & 0.798$\pm$0.003 & \underline{0.800$\pm$0.004} & \textbf{0.802$\pm$0.003}\\
80   & 0.816$\pm$0.003 & 0.829$\pm$0.003 & \underline{0.835$\pm$0.001} & \textbf{0.836$\pm$0.002}\\
70   & 0.839$\pm$0.002 & 0.851$\pm$0.002 & \underline{0.853$\pm$0.002} & \textbf{0.860$\pm$0.003}\\
\bottomrule
\end{tabular}}
\label{missing}}
\end{table*}

% \textcolor{red}{To validate the robustness of DrugBAN with high fraction of missing data, we conduct experiments with only a small number of DTI data for training. We train our model and each deep learning baseline with only 5\%, 10\%, 20\% and 30\% of one dataset, and predict on the rest of data (90\% as test set for evaluation and 10\% as validation set for early stopping). Table \ref{missing} presents the performance comparison under the missing data setting. We see that DrugBAN has the best performance in all settings. In particular, the improvement is more obvious on the larger dataset (BindingDB). As a result, our model demonstrates the greatest robustness to datasets the high fraction of missing data.}

We conduct experiments to clarify how the proposed model performs with high fraction of missing data on BindingDB and BioSNAP. Following the missing data setting in MolTrans \cite{Huang2021MolTransMI}, we train DrugBAN and deep learning baselines with only 5\%, 10\%, 20\% and 30\% of each dataset, and evaluate predictive performance on the rest of data (90\% as test set and 10\% as validation set for determining early stopping). Table \ref{missing} presents the obtained results, showing DrugBAN has the best performance in all settings. In particular, the improvement is larger on the bigger dataset (BindingDB).

\subsection*{S7 Scalability}
\label{sec:scalability}

\begin{figure}[hbt!]
\centering
    \includegraphics[width=1\textwidth]{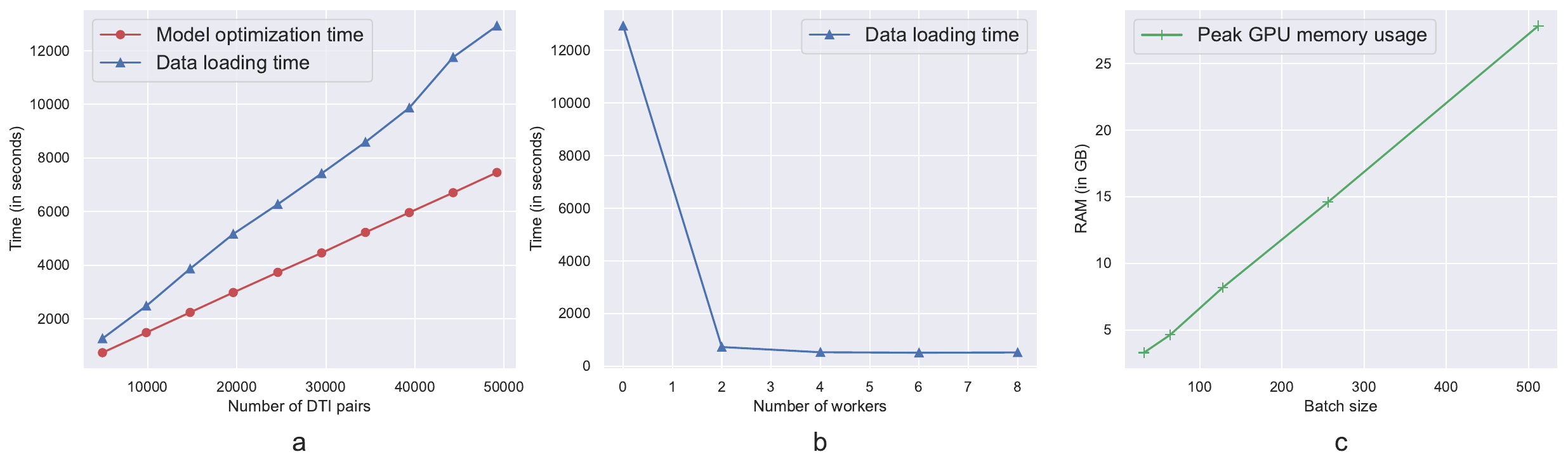}
    \caption{\textbf{Scalability of DrugBAN on the BindingDB dataset} (\textbf{a}) Model optimization and data loading time increase almost linearly with the number of DTI pairs. (\textbf{b}) Data loading time significantly reduces with the increasing number of workers. (\textbf{c}) Peak GPU memory usage increases linearly with the batch size.}
    \label{fig:scalability}
\end{figure}

We study the scalability of DrugBAN from three different perspectives: model optimization time, data loading time and GPU memory usage. We use the default hyperparameter configuration in Table \ref{Hyper}, and a single Nvidia V100 GPU to train the model in 100 epochs. Figure \ref{fig:scalability}a illustrates the model optimization time and data loading time against the number of DTI pairs for 4,919 (10\%) - 49,199 (100\%) from the BindingDB dataset. We empirically observe that the optimization time (red line) of DrugBAN increases almost linearly with the number of DTI pairs. It takes about two hours for 49,199 DTI pairs to complete the optimization. The data loading process (blue line) takes more time than model optimization. Nevertheless, since the data loading can be done on CPU, we can accelerate the process with multiple loading workers (subprocesses) in parallel. Figure \ref{fig:scalability}b shows the data loading time changes with respect to the number of workers, and it reduces significantly with only two additional workers added. Figure \ref{fig:scalability}c shows the peak GPU memory usage against the batch size. We find that DrugBAN only takes up 4.63 GB RAM with the default batch size 64, which is highly efficient. Similar to the optimization time, the memory usage also increases linearly with the batch size. This study demonstrates the scalability of DrugBAN.

\end{document}